\def\year{2020}\relax
\documentclass[letterpaper]{article} 
\usepackage{aaai20}  
\usepackage{times}  
\usepackage{helvet} 
\usepackage{courier}  
\usepackage[hyphens]{url}  
\usepackage{graphicx} 
\usepackage{graphicx}  
\usepackage[algoruled,linesnumbered]{algorithm2e}
\usepackage{amsmath}
\DeclareMathOperator*{\argmax}{arg\,max}

\newcommand{\psif}{\textsc{P-SIF}}
\usepackage{verbatim}

\usepackage[bottom,flushmargin,hang,multiple,para]{footmisc}

\title{\psif{}: Document Embeddings Using Partition Averaging}

\author{
Vivek Gupta\textsuperscript{\rm 1,3},
Ankit Saw\textsuperscript{\rm 2},
Pegah Nokhiz\textsuperscript{\rm 1},
Praneeth Netrapalli\textsuperscript{\rm 3},\\
\bf \Large
Piyush Rai\textsuperscript{\rm 4},
Partha Talukdar\textsuperscript{\rm 5}\\
\textsuperscript{\rm 1}School of Computing, University of Utah,
\textsuperscript{\rm 2}Info Edge (India) Limited,
\textsuperscript{\rm 3}Microsoft Research Lab, Bangalore,\\
\textsuperscript{\rm 4}Computer Science Department, IIT Kanpur,
\textsuperscript{\rm 5}Indian Institute of Science, Bangalore\\
vgupta@cs.utah.edu, ankit.kgpian@gmail.com, pnokhiz@cs.utah.edu,  \\ praneeth@microsoft.com, piyush@cse.iitk.ac.in, ppt@iisc.ac.in \\
}
\begin{document}
\maketitle

\begin{abstract}
Simple weighted averaging of word vectors often yields effective representations for sentences which outperform sophisticated seq2seq neural models in many tasks. While it is desirable to use the same method to represent documents as well, unfortunately, the effectiveness is lost when representing long documents involving multiple sentences. One of the key reasons is that a longer document is likely to contain words from many different topics; hence, creating a single vector while ignoring all the topical structure is unlikely to yield an effective document representation. This problem is less acute in single sentences and other short text fragments where the presence of a single topic is most likely. To alleviate this problem, we present \psif{}, a partitioned word averaging model to represent long documents. \psif{} retains the simplicity of simple weighted word averaging while taking a document's topical structure into account. In particular, \psif{} learns topic-specific vectors from a document and finally concatenates them all to represent the overall document. We provide theoretical justifications on the correctness of \psif{}. Through a comprehensive set of experiments, we demonstrate \psif{}'s effectiveness compared to simple weighted averaging and many other baselines. 
\end{abstract}

\section{Introduction}
Many approaches such as \cite{socher2013recursive,liu2015learning,le2014distributed,wang2015} are proposed which go beyond words to capture the semantic meaning of sentences. These techniques either use the simple composition of the word-vectors or sophisticated neural network architectures for sentence representation. Recently, \cite{arora2016simple} proposed a smooth inverse frequency (\textsc{SIF}) based word vector averaging model to embed a sentence. They further improved their embedding by removing the first principal component of the weighted average vectors. However, all these approaches are limited to capturing the meaning of a single sentence and representing the sentence in the same space as words, thus reducing their expressive power. Generally, longer texts contain words from multiple topics, so creating a single vector from simple averaging of word-vectors will disregard all the topical structure. \footnote{Topical structure denotes word distribution across topics.} Hence, these techniques are largely unable to capture the semantic meanings of larger pieces of text, e.g., multi-sentence documents.  

To address these limitations, we present a novel document embedding method called  \emph{partition SIF} weighted averaging (\psif{}) to embed documents which usually contain multiple sentences efficiently. \psif{} learns topic-specific vectors from a document and finally concatenates them all to represent the overall document. Thus, \psif{} retains the simplicity of simple weighted word averaging while taking a document's topical structure into account. We also provide theoretical justifications for the proposed approach and demonstrate its efficacy via a comprehensive set of experiments. \psif{} achieves significant improvements over several embedding techniques on several tasks despite being simple. We have released the source code for \psif{} embeddings. \footnote{\url{https://github.com/vgupta123/P-SIF}} The novel characteristics of \psif{} are described below:
\begin{itemize}
\itemsep 0.04em 
\item \psif{} can embed larger multi-sentence documents, as it pays attention to the topical structure of the document.
\item \psif{} is based on simple weighted word vectors averaging rather than considerably more sophisticated tensor factorization or neural network-based methods.
\item \psif{} is unsupervised since it only uses pre-trained word embeddings without using any label information.
\item \psif{} outperforms many existing methods on text similarity, text classification, and other supervised tasks.
\end{itemize} 

\section{Related Work}
\label{sec:relatedwork}
Most of the prior work has computed sentence embeddings by coordinate wise vector and matrix-based compositional operations over word vectors, e.g., \cite{levy2014neural} use unweighted averaging of word vectors\cite{le2014distributed} for representing short phrases, \cite{pranjal2015weighted} propose tfidf-weighted averaging of word vectors to form document vectors, \cite{socher2013recursive} propose a recursive neural network defined over a parse tree, and trained with supervision. 

Next, \cite{le2014distributed} propose \textit{PV-DM} and \textit{PV-DBOW} models which treat each sentence as a shared global latent vector. Other approaches use seq2seq models such as Recurrent Neural Networks \cite{mikolov2010recurrent} and Long Short Term Memory \cite{gers2002learning} which can handle long term dependency, hence capturing the syntax structure. Other neural network models include the use of hierarchy and convolutional neural networks such as \cite{kim2014convolutional}. \cite{wieting2015paraphrase} learns paraphrastic sentence embeddings by modifying word embeddings via supervision from the Paraphrase pairs dataset (PPDB) \cite{ganitkevitch2013ppdb}. 

Recently, a lot of work is harnessing topic modeling \cite{Blei:2003} along with word vectors to learn better word and sentence representations, e.g., LDA \cite{liu2014topic}, weight-BoC \cite{boc}, TWE  \cite{AAAI159314} , NTSG \cite{liu2015learning}, WTM \cite{wtm}, w2v-LDA \cite{wtvlda}, TV+MeanWV \cite{tvMeanWV}, LTSG \cite{ltsg}, Gaussian-LDA \cite{gaussianlda}, Topic2Vec \cite{topic2vec}, TM \cite{dieng2019topic}, LDA2vec \cite{lda2vec}, D-ETM \cite{dieng2019dynamic} and MvTM \cite{mvtm}. \cite{kiros2015} propose skip-thought document embedding vectors which transformed the idea of abstracting the distributional hypothesis from word to sentence level. \cite{wieting2015towards} propose a neural network model which optimizes the word embeddings based on the cosine similarity of the sentence embeddings. Moreover, several recent deep contextual word embeddings such as ELMo \cite{Peters:2018}, USE \cite{cer2018universal} and BERT \cite{devlin2018bert} are proposed. These contextual embeddings are state-of-the-art on multiple tasks as they effectively capture the surrounding contexts. 
  
\cite{vivek} propose methods which employ a clustering-based technique and tf-idf values to form a composite document vector extending the Bag-of-Words (BoW) model \cite{harris54}. They represent documents in higher dimensions by using hard clustering over word embeddings. \cite{mekala2017scdv} extend this by proposing SCDV using an overlapping clustering technique and direct idf weighting of word vectors. The learned representations try to capture a global context of a sentence, similar to an $n$-gram model. Our method is the same in essence, but is based on topic-based partitioning; moreover, unlike \cite{mekala2017scdv}'s approach, our method is supported by theoretical guarantees. 

\section{Averaging vs Partition Averaging}
\label{sec:motivation}
\begin{figure}[ht]
\centering
\includegraphics[width=.96\columnwidth]{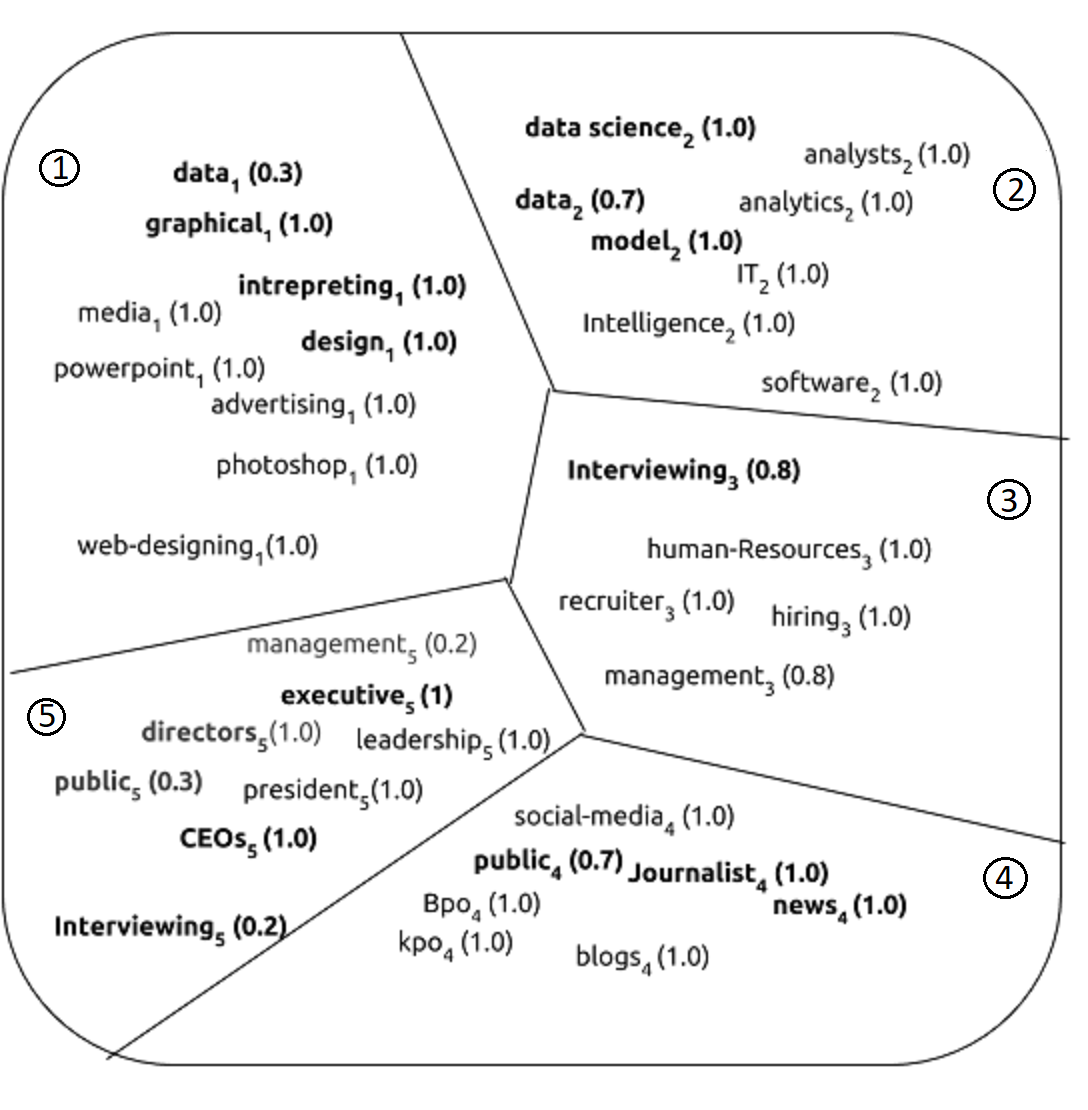}
\caption{Words in different partitions are represented by different subscripts and separated by hyper-planes. Bold fonts represent words' presence in document $d_n$.}
\label{figure:illustration}
\end{figure}

Figure \ref{figure:illustration}, represents the word-vector space, where similar meaning words occur closer to each other. We can apply sparse coding to partition the word-vector space to a five topic vector space. These five topic vector spaces represent the five topics present in corpus. Some words are multi-sense and belong to multiple topics with some proportion. In Figure \ref{figure:illustration} we represent words' topic number in subscript and proportion in brackets. Let's consider a document $d_n$: \textit{``Data journalists deliver data science news to general public. They often take part in interpreting the data models. In addition, they create graphical designs and interview the directors and CEOs."}

If we directly average words to represent document ($\vec{v}_{d_{n}}$), as is done in SIF~\cite{arora2016simple}, then different semantic meaning words, e.g., words in partition $1$ such as \textit{`graphical'}, \textit{`design'}, and \textit{`data'} are averaged with words of different semantic meaning of partition $2$ such as  \textit{`data science'}, \textit{`model'}, and \textit{`data'}. In addition, the document is represented in the same $d$ dimensional space as word vectors. Overall, averaging represents the documents as a single point in the vector space and does not consider the \textit{5} different semantic topics. However, we can weight (topic proportion) average of words within a partition and concatenate ($\oplus$) the average word vectors across partitions to represent document ($\vec{v}_{d_{n}}$), as is done in our proposed method \psif{}. By doing this, words belonging to different semantic topics are separated by concatenation ($\oplus$) as they represent separate meanings, whereas words in similar topics are simply averaged since they represent the same meaning. For example, average of words belonging to partition $1$ such as \textit{`graphical'}, \textit{`design'}, and \textit{`data'} are concatenated to average of words in partition $2$ such as  \textit{`data science'}, \textit{`model'}, and \textit{`data'}. The final document vector $\vec{v}_{d_{n}}$ is represented in a higher $5$ $\times$ $d$ dimension vector space, thus having more representational power ($d$ is the dimension of word vectors).  Overall, the $5$ different semantic topics are taken into account for representation. Additionally, this representation also takes the weight according to which each word belongs to various topics into account, meaning it handles words' multi-sense natures. For example, \textit{`data'} belongs to partition $1$ with probability $0.3$ and partition $2$ with probability $0.7$. Hence, partitioned averaging with topic weighting is essential for representing longer text documents.

\section{The Proposed Algorithm: \psif{}}
\label{sec:algorithm}
In this section, we present the new proposed document embedding learning method in algorithm \ref{algo:SIF_MODIFIED}. The feature formation algorithm can be divided into three major steps:

\textbf{Sparse Dictionary Learning for Word Vectors (Algo \ref{algo:SIF_MODIFIED}: Lines \ref{algo:dlstarting} - \ref{algo:dlend}): } Given word vectors $v_{w} \in R^d$, a sparsity parameter $k$, and an upper bound $K$ , we find a set of unit norm vectors $\vec{A}_1, \vec{A}_2, \ldots , \vec{A}_K$, such that {\small $\vec{v}_{w} =  \sum_{j=1}^{K} \alpha_{(w,j)}\vec{A}_{j} + \vec{\eta}_{w}$ } where at most $k$ out of $K$ of the coefficients $\alpha_{(w,1)}, \ldots ,\alpha_{(w,K)}$ are nonzero (so-called sparsity constraint), and $\vec{\eta}_{w}$ is a noise vector. Sparse coding is usually solved for a given $K$ and $k$ by using alternating minimization such as k-svd \cite{aharon2006k} to find the $\vec{A}_{i}'$s that minimize the following $L_2$-reconstruction error: {\small $\| \vec{v}_{w} - \sum_{j=1}^{K} \alpha_{(w,j)}\vec{A}_{j} \|$}. \cite{arora2016linear} show that  multiple senses of a word reside as a linear superposition within the word embedding and can be recovered by simple sparse coding. Therefore, one can use the sparse coding of word vectors to detect multiple senses of words. Additionally, the atoms of sparse coding ($\vec{A}_{1}, \ldots, \vec{A}_{K}$) over word-vectors ($\vec{v}_{w}$) represent all prominent topics in the corpus. For a given word $w$, the $k$ non-zero coefficient of $\alpha_w$ essentially represents the distribution of words over topics. Furthermore, restricting $K$ to be much smaller than the number of the words ensures that the same topic needs to be used for multiple words. The learned $\vec{A}_j$ is a significant topic because the sparse coding ensures that each basis element is softly chosen by many words. 

\textit{Sparse Dictionary Learning vs. Overlapping Clustering: } Sparse coding can also be treated as a linear algebraic analogue of overlapping clustering, where the $\vec{A}_{i}$'s act as cluster centers and each $\vec{v}_{w}$ is assigned to each cluster in a soft way (using the coefficients $\alpha_{(w,j)}$, of which only $k$ out of $K$ are non-zero) to a linear combination of at most $k$ clusters. In practice, sparse coding optimization produces coefficients $\alpha_{(w,j)}$ which are almost all positive, even though \textit{unconstrained}. One can use overlapping clustering where each word belongs to every cluster with some probability $P(c_{k}|w _{i})$ which can be thought of as a substitute for $\alpha_{(w,k)}$, similar to the approach in SCDV \cite{mekala2017scdv}. Instead of GMM, we use a dictionary learning-based approach which imposes a sparsity constraint implicitly during optimization through regularization. Additionally, such high dimensional data structure regularizers, e.g., sparse encodings, help in overcoming the curse of high dimensionality. For single-sentence documents with a small number of topics, it is better to use overlapping clustering because of an easier unconstrained optimization. However, in case of multi-sentence documents where the number of topics is large, dictionary learning performs better than overlapping clustering due to 1) Sparse constraint optimization forces non-redundant clusters (minimally sufficient $\#$clusters) and 2) The sparse constraint diminishes the noise from the long tail of word-cluster assignments $P(c_{k}|w _{i})$ \cite{olshausen1997sparse,gao2010local,yang2009linear}. 

\textbf{Word Topics Vector Formation (Algo \ref{algo:SIF_MODIFIED}: Lines \ref{algo:wtvstart} - \ref{algo:wtvend}): } For single sentence documents all words of a document belong to a single topic. However, for multi-sentence documents, words of a document generally originate from multiple topics. To capture this, topic modeling algorithms such as LDA \cite{Blei:2003} are used to represent the documents. These representations essentially represent the global contexts of the documents as a distribution over topics. However, these representations do not take the local context initiating from the distributional semantics such as word vectors into account.
Since our multi-sentence documents have words from multiple topics, a simple averaging technique will not work. Hence, we concatenate the word embeddings over words' topic distributions. This helps to represent semantically similar words in the same topic, while words which are semantically different are represented in different topics. Concatenation of word embeddings over topics also helps in the expression of words' multi-sense nature.
For each word $\vec{w}$, we create $K$ different word-cluster vectors of $d$ dimensions $\vec{cv}_{wk}$ by weighting the word embedding with its learned dictionary coefficient $\alpha_{w,k}$ of the $k^{th}$ context. \footnote{Empirically, we observed that this weighting generally improves the performance.} We then concatenate all the $K$ word-cluster vectors $\vec{cv}_{wk}$ into a $K \times d$ dimensional embedding to form a word-topic vector $\vec{tv}_w$ $\in$ $R^{K \times d}$. We weigh word-vectors by coefficients of the learned dictionary to capture the cross correlation $({\alpha_i}{\alpha_j})$ between topics. The word-topic-vector $\vec{tv_w}$, which we average to represent documents, captures both local and global semantics. 

\begin{algorithm}[ht]
    \SetAlgoNoLine
    \KwData{$d$ dimensional Word embeddings $\{\vec{v}_{w}: w \in V\}$ where word $w$ is in vocabulary $V$. Documents $\{d_n : d_n \in D\}$, a set of sentences $D$ in corpus $C$, parameter $a$ and estimated unigram probability $\{p(w) : w \in V\}$ of word $w$ in $C$, a sparsity parameter $k$, and an upper bound $K$.}
    \KwResult{Document vectors $\{{\vec{v}}_{d_n} : d_n \in D\}$}
    \tcc{Dictionary learning on word-vectors}    
    \For{each word $w$ $in$ $V$}{
        \label{algo:dlstarting}
      $\vec{v}_{w}$ $=$ $\sum_{j=1}^{K}$ $\alpha_{w,j}\vec{A}_{j}$ $+$ $\vec{\eta}_{w}$\;
        }
    \label{algo:dlend}
    \tcc{Word topic-vector formation}
    \For{each word $w$ $in$ $V$}{
        \label{algo:wtvstart}
        \For{each coefficient, $\alpha_{w,k}$ of word $w$}{
             $\vec{cv}_{w,k}$ $\leftarrow$ $\vec{v}_w$ $\times$  $\alpha_{w,k}$\;
        }
        $\vec{tv}_{w}$ $\leftarrow$ $\bigoplus_{k=1}^K$ $\vec{cv}_{wk}$ \; \tcc{$\bigoplus$ is concatenation, $\times$ is scalar vector multiplication}
    }
    \label{algo:wtvend}
    \tcc{ SIF reweighed embedding } 
     \For{each document $d_n$ in $D$}{
         \label{algo:sifstart}
            ${\vec{v}}_{d_n}$ $\leftarrow$ $\frac{1}{|d_n|}$ $\sum_{w \in d_n} \frac{a}{a +p(w)}$ $\vec{tv}_{w}$\;
    }
   Form a matrix $X$ whose columns are $\{\vec{v}_{d_n} : d_n \in D\}$, and let $\vec{u}$ be the first singular vector\;
   \For{each document $d_n$ $\in$ $D$}{
            ${\vec{v}}_{d_n}$ $\leftarrow$ ${\vec{v}}_{d_n}$ - $\vec{u}\vec{u}^{T}{\vec{v}}_{d_n}$ \;
    }
   \label{algo:sifend}
\caption{\psif{} Embedding}
\label{algo:SIF_MODIFIED}
\end{algorithm}

\textbf{SIF Weight Averaging and Common Component Removal (Algo \ref{algo:SIF_MODIFIED}: Lines \ref{algo:sifstart} - \ref{algo:sifend}): } Finally, for all words appearing in document $D_n$, we weight the word-topics vectors $\vec{tv}_i$ by smooth inverse frequency \big($\frac{a}{a + p(w)}$\big). Next, we remove the common contexts from the weighted average document vectors by removing the first principal component from the weighted average vectors. \footnote{We did not remove the common component from final vectors when we used Doc2VecC-initialized \cite{chen2017efficient} word vectors with \psif{}. Because frequent words' word-vectors become close to $\vec{0}$.} Common component removal reduces the noise and redundancy from the document vectors which makes the representations more discriminating. \cite{arora2016simple} empirically shows that SIF weighting outperforms the tf-idf weighting. However, they only use simple averaging to represent a sentence. Detailed code architecture of \psif{} is in the supplementary material. \footnote{\url{https://vgupta123.github.io/docs/appendix_aaai2020.pdf}}

\textbf{Derivation of \psif{} Embeddings} : We provide theoretical justifications by showing connections of \psif{} with random-walk based latent variable models~\cite{arora2016simple,arora2016latent,arora2016linear}. Full derivations are provided in the supplementary material $^5$.

\section{Kernels meet Embeddings}
In this section, we present one of the novelties of this work where we show that many common sentence embeddings can be represented as similarity kernels over word and topic vectors. Let  $D_A$ and $D_B$ represent two documents containing $n$ and $m$ words respectively. ${w^{A}_1,w^{A}_2 \ldots w^{A}_n}$ denotes $D_A$'s words and ${w^{B}_1,w^{B}_2 \ldots w^{B}_m}$  denotes $D_B$'s words. 

\begin{enumerate}
\item Simple Word Vector Averaging : {\small $K^1(D_A,D_B) =  \frac{1}{nm}\sum_{i=1}^{n} \sum_{j=1}^{m} \langle\vec{v}_{w^{A}_{i}}\cdot \vec{v}_{w^{B}_{j}} \rangle $}

\item TWE: Topical Word  Embeddings \cite{AAAI159314} : {\small $K^2(D_A,D_B) = \frac{1}{nm} \sum_{i=1}^{n} \sum_{j=1}^{m} \langle \vec{v}_{w^{A}_{i}} \cdot \vec{v}_{w^{B}_{j}}\rangle + \langle\vec{tv}_{w^{A}_{i}} \cdot \vec{t}_{w^{B}_{j}}\rangle $} 

\item \psif{}: Partition Word Vector Averaging (Our approach) : {\small $K^3(D_A,D_B) = \frac{1}{nm} \sum_{i=1}^{n} \sum_{j=1}^{m} \langle \vec{v}_{w^{A}_{i}} \cdot \vec{v}_{w^{B}_{j}}\rangle \times \langle\vec{t}_{w^{A}_{i}} \cdot \vec{t}_{w^{B}_{j}}\rangle$}

\item Relaxed Word Mover Distance \cite{kusner2015word} : {\small $K^4(D_A,D_B) = \frac{1}{n} \sum_{i=1}^{n} \max_{j} \langle\vec{v}_{w^{A}_{i}} \cdot \vec{v}_{w^{B}_{j}}\rangle$}

\end{enumerate}

Here, $\vec{v}_{w}$ represents the word vector of word $w$ and $\vec{t}_{w} = \alpha_{w}$ $\in$ $R^K$ represents the topic vector of word $w$, where $K$ is the number of topics. $\langle \vec{a} \cdot \vec{b} \rangle$ represents the dot product of two vectors $\vec{a}$ and $\vec{b}$. $c \times d$ represents the scalar product of $c$ and $d$. $\bigoplus$ represents the row-wise concatenation of the vectors. Refer to the Supplementary material $^5$ for the detailed proof.  

\section{Experimental Results}
We perform a comprehensive set of experiments on several text similarity and multiclass or multilabel text classification tasks. Due to limited space, some details on experiments are in the Supplementary material $^5$.

\label{sec:experiment}
\subsection{Textual Similarity Task}
\label{sec:unsupervised}
\textbf{Datasets and Baselines: } We perform our experiments on the SemEval dataset (2012 - 2017). These experiments involve $27$ semantic textual similarity (STS) tasks (2012 - 2016) \cite{agirre2012semeval,agirre2016semeval} the SemEval 2015 Twitter task \cite{xu2015semeval}, and the SemEval 2014 Semantic relatedness task \cite{marelli2014semeval}. The objectives of these tasks are to predict the similarity between two sentences. We compare our approach with several unsupervised, semi-supervised and supervised embedding baselines mostly taken from \cite{arora2016simple,D18-1482,ethayarajh2018unsupervised}. Details on the baselines are listed below: 

\textbf{Unsupervised}: We used ST, avg-GloVe, tfidf-GloVe, and GloVe + WR as a baseline. ST denotes the skip-thought vectors by \cite{kiros2015}, avg-GloVe denotes the unweighted average of the GloVe Vectors by \cite{pennington2014glove} \footnote{We used the 300-dimensional word vectors that are publicly available at \url{http://nlp.stanford.edu/projects/glove/}.}, and tfidf-Glove denotes the tf-idf weighted average of GloVe vectors. We also compared our method with the SIF weighting ($W$) common component removal ($R$) GloVe vectors (GloVe + $WR$) by \cite{arora2016simple}. 
For STS 16, we also compared our embedding with Skip-Thoughts \cite{kiros2015}, BERT pretrained embedding average \cite{devlin2018bert} , Universal Sentence Encoder \cite{cer2018universal} and Sent2Vec \cite{pgj2017unsup} embeddings.

\textbf{Semi-Supervised}: We used avg-PSL, PSL + WR, and the avg-PSL used the unweighted average of the PARAGRAM-SL999 (PSL) word vectors by \cite{wieting2015paraphrase} as a baseline, obtained by training on PPDB dataset\cite{ganitkevitch2013ppdb}. The word vectors are trained using unlabeled data. Furthermore, sentence embeddings are obtained from unweighted word vectors averaging. We also compared our method with the SIF weighting (W) common component removal (R) PSL word vectors (PSL + WR) by \cite{arora2016simple}.

\textbf{Supervised}: We compared our method with PP, PP-proj., DAN, RNN, iRNN, LSTM (o.g), LSTM (no) and GRAN. All these methods are initialized with PSL word vectors and then trained on the PPDB dataset~\cite{ganitkevitch2013ppdb}. PP~\cite{wieting2015towards} is the average of word vectors, while PP-proj is the average of word vectors followed by a linear projection. The word vectors are updated during the training. DAN denotes the deep averaging network~\cite{iyyer2015deep}. RNN is a Recurrent neural network, iRNN is the identity activated Recurrent Neural Network based on identity-initialized weight matrices. The LSTM is the version from \cite{gers2002learning}, either with output gates (denoted as LSTM (o.g.)) or without (denoted as LSTM (no)). GRAN denotes state of the art supervised averaging based Gated Recurrent Averaging Network from \cite{wieting2017revisiting}. For STS 16 we also compared our embedding with Tree-LSTM \cite{tai2015improved} embedding.

\textbf{Experimental Settings: }We use the Pearson's coefficient between the predicted and the ground-truth scores for the evaluation. We use the {PARAGRAM-SL999 (PSL)} from \cite{wieting2015paraphrase} as word embeddings, obtained by training on the PPDB \cite{ganitkevitch2013ppdb} dataset \footnote{For a fair comparison with SIF  we use PSL vectors instead of unsupervised GloVe and Word2Vec vectors.}. We use the fixed weighting parameter $a$ value of $10^{-3}$, and the word frequencies $p(w)$ are estimated from the common-crawl dataset. We tune the number of contexts $(K)$ to minimize the reconstruction loss over the word-vectors. We fix the non-zero coefficient $k$ = $K/2$, for the SIF experiments. For the GMM-based partitioning of the vocabulary, we tune the number of clusters' parameter $K$ through a 5-fold cross validation.

\begin{table*}[ht]
\begin{center}
\small
\caption{Experimental results (Pearson's r $\times$ 100) on textual similarity tasks. Many results are collected from \cite{wieting2015towards}, DAN \cite{iyyer2015deep} and \cite{wieting2017revisiting} (GRAN) except for tfidf-GloVe.}
\label{table:3}
\vspace{1.0em}
\resizebox{1.98\columnwidth}{!}{
\begin{tabular}{ c|c|c|c|c|c|c|c|c|c|c|c|c|c|c|c } 
\hline
\bf Tasks & \bf PP & \bf PP & \bf DAN & \bf RNN & \bf iRNN & \bf LSTM & \bf LSTM & \bf GRAN &\bf ST & \bf Avg & \bf tfidf &\bf  Avg & \bf Glove & \bf PSL & \bf PSIF  \\
& &\bf proj & & & & \bf (no) &\bf (o.g.) & & & \bf Glove & \bf Glove & \bf PSL &\bf +WR &\bf +WR &\bf +PSL \\\hline
STS’12 & 58.7 & 60.0 & 56.0 & 48.1 & 58.4 & 51.0 & 46.4 & 62.5 & 30.8 & 52.5 & 58.7 & 52.8 & 56.2 & 59.5 &\bf 65.7\\ 
STS’13 & 55.8 & 56.8 & 54.2 & 44.7 & 56.7 & 45.2 & 41.5 & 63.4 & 24.8 & 42.3 & 52.1 & 46.4 & 56.6 & 61.8 &\bf 64.0\\ 
STS’14 & 70.9 & 71.3 & 69.5 & 57.7 & 70.9 & 59.8 & 51.5 &\bf 75.9 & 31.4 & 54.2 & 63.8 & 59.5 & 68.5 & 73.5 & 74.8\\
STS’15 & 75.8 & 74.8 & 72.7 & 57.2 & 75.6 & 63.9 & 56.0 &\bf 77.7 & 31.0 & 52.7 & 60.6 & 60.0 & 71.7 & 76.3 & 77.3\\
Sick’14 & 71.6 & 71.6 & 70.7 & 61.2 & 71.2 & 63.9 & 59.0 & 72.9 & 49.8 & 65.9 & 69.4 & 66.4 & 72.2 & 72.9 &\bf 73.4\\
Twit15 & 52.9 & 52.8 & 53.7 & 45.1 & 52.9 & 47.6 & 36.1 & 50.2 &24.7 & 30.3 & 33.8 & 36.3 & 48.0 & 49.0 &\bf 54.9\\
\hline
\end{tabular}}
\end{center}
\end{table*}
\begin{table*}[t]
\small
\centering
\caption{P-SIF comparison with the recent embedding techniques on various STS tasks. Baselines taken from \cite{conneau2018senteval}, \cite{perone2018evaluation}, \cite{cer2018universal}, \cite{devlin2018bert}, \cite{D18-1482} and \cite{ethayarajh2018unsupervised}}
\vspace{1.0em}
\resizebox{1.98\columnwidth}{!}{
\begin{tabular}{c|c|c|c|c|c|c|c|c|c|c|c|c}
\hline
\bf Task &\bf ELMo &\bf ELMo & \bf Bert(pr) &\bf USE &\bf p-mean &\bf Fast &\bf Skip & \bf Infer &\bf Char &\bf WME  &\bf PSIF &\bf u-SIF\\ 
 &\bf orig+all &\bf orig+top & \bf Avg. &  &  &\bf Text  &\bf Thoughts &\bf Sent &\bf pharse &\bf  +PSL &\bf  +PSL &\bf  +PSL\\ \hline
STS 12 & 55 & 54 & 53 & 65 & 54 & 58 & 41 & 61 & 66 & 62.8 & 65.7 & \bf 65.8\\ 
STS 13 & 51 & 49 & 67 & 68 & 52 & 58 & 29 & 56 & 57 & 56.3 & 63.98 &\bf 65.2\\ 
STS 14 & 63 & 62 & 62 & 64 & 63 & 65 & 40 & 68 & 74.7 & 68.0 & 74.8 &\bf 75.9\\ 
STS 15 & 69 & 67 & 73 & 77 & 66 & 68 & 46 & 71 & 76.1 & 64.2 & 77.3 &\bf 77.6\\ 
STS 16 & 64 & 63 & 67 & 73& 67 & 64 & 52 & 77 & - & - &\bf 73.7 & 72.3\\ \hline
Average & 60.4 & 59 & 64.4 & 69.4 & 60.4 & 62.6 & 41.6 & 66.6 & 68.5 & 62.9 & 71.1 &\bf 71.4\\ \hline
\end{tabular}}
\label{tab:sts_more}
\end{table*}

\begin{table*}[!htbp]
\small
\centering
\caption{Comparison of \psif{} (SGNS) with the recently proposed word mover distance and word mover embedding \cite{D18-1482} based on accuracy. In $\pm x$, $x$ is the variance across several runs. }
\vspace{1.0em}
\resizebox{1.95\columnwidth}{!}{
\begin{tabular}{c|c|c|c|c|c|c|c|c}
\hline
\bf Dataset & \bf Bbcsport & \bf Twitter &\bf Ohsumed &\bf Classic & \bf Reuters &\bf Amazon &\bf 20NG &\bf Recipe-L \\ \hline
SIF(GloVe) & 97.3 $\pm$ 1.2 & 57.8 $\pm$ 2.5 &\bf 67.1 & 92.7 $\pm$ 0.9 & 87.6 & 94.1 $\pm$ 0.2 & 72.3 & 71.1 $\pm$ 0.5  \\ 
Word2Vec Avg  & 97.3 $\pm$ 0.9 & 72.0 $\pm$ 1.5 & 63 & 95.2 $\pm$ 0.4 & 96.9 & 94.0 $\pm$ 0.5 & 71.7 & 74.9 $\pm$ 0.5  \\ 
PV-DBOW & 97.2 $\pm$ 0.7 & 67.8 $\pm$ 0.4 & 55.9 & 97.0 $\pm$ 0.3 & 96.3 & 89.2 $\pm$ 0.3 & 71 & 73.1 $\pm$ 0.5  \\ 
PV-DM & 97.9 $\pm$ 1.3 & 67.3 $\pm$ 0.3 & 59.8 & 96.5 $\pm$ 0.7 & 94.9 & 88.6 $\pm$ 0.4 & 74 & 71.1 $\pm$ 0.4  \\ 
Doc2VecC & 90.5 $\pm$ 1.7 & 71.0 $\pm$ 0.4 & 63.4 & 96.6 $\pm$ 0.4 & 96.5 & 91.2 $\pm$ 0.5 & 78.2 & 76.1 $\pm$ 0.4  \\
KNN-WMD & 95.4  $\pm$  1.2 & 71.3  $\pm$  0.6 & 55.5 & 97.2  $\pm$  0.1 & 96.5 & 92.6  $\pm$  0.3 & 73.2 & 71.4  $\pm$  0.5  \\
SCDV & 98.1 $\pm$ 0.6 & 74.2 $\pm$ 0.4 & 53.5 & 96.9 $\pm$ 0.1 & 97.3 & 93.9 $\pm$ 0.4 & 78.8 & 78.5$\pm$ 0.5 \\
WME & 98.2 $\pm$ 0.6 &\bf 74.5 $\pm$ 0.5 & 64.5 & 97.1 $\pm$ 0.4 & 97.2 & 94.3 $\pm$ 0.4 & 78.3 &\bf 79.2 $\pm$ 0.3  \\  \hline
P-SIF & 99.05 $\pm$ 0.9 & 73.39 $\pm$ 0.9 &\bf 67.1 & 96.95 $\pm$ 0.5 &\bf 97.67 & 94.17 $\pm$ 0.3 & 79.15 & 78.24 $\pm$ 0.3 \\
P-SIF (Doc2VecC)&\bf 99.68 $\pm$ 0.9 & 72.39 $\pm$ 0.9 &\bf 67.1 &\bf 97.7 $\pm$ 0.5 & 97.62 &\bf 94.83 $\pm$ 0.3 &\bf 86.31 & 77.61 $\pm$ 0.3 \\ \hline
\end{tabular}}
\label{tab:my_wnd_comp}
\end{table*}

\textbf{Results and Analysis: }The average results for each year are reported in Tables \ref{table:3} and \ref{tab:sts_more}.  We denote our embeddings by \psif{} $+$ PSL ($+$ PSL denotes using the PSL word vectors). We report the average results for the STS tasks. The detailed results on each sub-dataset are in the Supplementary material $^5$. We observe that  \psif{} + PSL outperforms PSL + WR on all datasets, thus supporting the usefulness of our partitioned averaging. Despite being simple, our method outperforms many complicated methods such as seq2seq, Tree-LSTM\cite{tai2015improved}, and Skip-Thoughts\cite{kiros2015}. We observe that partitioning through overlapping clustering algorithms such as GMM generates a better performance compared to partitioning through sparse dictionary algorithms such as k-svd for some Semantic Textual Similarity (STS) task datasets. The main reason for this peculiar observation is related to the fact that some STS datasets contain documents which are single sentences of a maximum length of $40$ words. As discussed earlier (sparse dictionary learning vs. overlapping clustering), for single sentence documents with a small number of topics, overlapping clustering optimizes better than sparse dictionary learning. Therefore, we use GMM for partitioning words into suitable clusters for some STS tasks. But both k-svd and GMM outperform simple averaging (SIF) by significant margins on most STS tasks. \footnote{k-svd always outperforms GMM on both classification datasets since the documents are multi-sentence with $\#$words $>>$ $40$.}  We also report qualitative results with real examples in the Supplementary $^5$.

\subsection{Text Classification Task}
\label{sec:textclassification}
The document embeddings obtained by our method can be used as direct features for many classification tasks.

\textbf{Datasets and Baselines: } We run multi-class experiments on 20NewsGroup dataset, \footnote{\url{https://bit.ly/2pqLCaN}} and multi-label classification experiments on Reuters-21578 dataset. \footnote{\url{https://goo.gl/NrOfu}} We use \textit{script} for preprocessing the dataset. \footnote{\url{https://gist.github.com/herrfz/7967781}} We consider several embedding baselines mostly taken from \cite{mekala2017scdv,D18-1482,arora2016linear}.  More details on experimental settings and hyper-parameters' values are described in the Supplementary material $^5$. We considered the following baselines: The Bag-of-Words (BoW) model \cite{harris54}, the Bag of  Word Vector (BoWV) \cite{vivek} model, Sparse Composite Document Vector (SCDV) \cite{mekala2017scdv} \footnote{\url{https://github.com/dheeraj7596/SCDV}} paragraph vector models \cite{le2014distributed}, Topical word  embeddings (TWE-1)  \cite{AAAI159314}, Neural Tensor Skip-Gram Model (NTSG-1 to NTSG-3) \cite{liu2015learning}, tf-idf weighted average word-vector model\cite{pranjal2015weighted} and weighted Bag of Concepts (weight-BoC) \cite{boc} where we built document-topic vectors by counting the member words in each topic, and Doc2VecC \cite{chen2017efficient} where averaging and training of word vectors are done jointly. Moreover, we used SIF \cite{arora2016simple} smooth inverse frequency weight with common component removal from weighted average vectors as a baseline. We also compared our results with other topic modeling based document embedding methods such as WTM \cite{wtm}, w2v-LDA \cite{wtvlda}, LDA \cite{liu2014topic}, TV+MeanWV \cite{tvMeanWV}), LTSG \cite{ltsg}, Gaussian-LDA  \cite{gaussianlda}, Topic2Vec \cite{topic2vec}, Lda2Vec \cite{lda2vec}, MvTM \cite{mvtm} and BERT \cite{devlin2018bert}. For BERT, we reported the results on the unsupervised pre-trained (pr) model because of a fair comparison to our approach which is also unsupervised.

\textbf{Experimental Settings:} We fix the document embeddings and only learn the classifier.
We learn word vector embeddings using Skip-Gram with a window size of $10$, Negative Sampling (SGNS) of $10$, and minimum word frequency of $20$. We use $5$-fold cross-validation on the $F1$ score to tune hyperparameters. We use LinearSVM for multi-class classification and Logistic regression with the OneVsRest setting for multi-label classification. We fix the number of dictionary elements to either $40$ or $20$ (with Doc2vecC initialize word vectors) and non-zero coefficient to $k$ = $K/2$ during dictionary learning for all experiments. We use the best parameter settings, as reported in all our baselines to generate their results. We use $200$ dimensions for tf-idf weighted word-vector model, $400$ for paragraph vector model, $80$ topics and $400$ dimensional vectors for TWE, NTSG, LTSG and $60$ topics and $200$ dimensional word vectors for SCDV \cite{mekala2017scdv}. We evaluate the classifiers' performance using standard metrics such as accuracy, macro-averaging precision, recall and F-score for multiclass classification tasks.We evaluate multi-label classifications' performance using Precision@K, nDCG@k, Coverage error, Label ranking average precision(LRAPS) and F1 score.\footnote{\url{https://goo.gl/4GrR3M}}

\begin{table}[ht]
\small
\caption{Multi-class classification performance on 20NewsGroups.}
\begin{center}
\begin{tabular}{ c|c|c|c|c } 
 \hline
 {\bf Model} & {\bf Acc} & {\bf Prec} & {\bf Rec} & {\bf Fmes} \\
 \hline
 \psif{} (Doc2VecC) & \bf {86.0} & \bf {86.1} &\bf {86.1}  &\bf {86.0} \\
 \psif{} & \bf {85.4} & \bf {85.5} &\bf {85.4}  &\bf {85.2} \\ \hline
 BERT (pr) & 84.9 & 84.9 & 85.0 & 85.0\\
 SCDV & 84.6 & 84.6 & 84.5  & 84.6 \\
 Doc2VecC & 84.0 & 84.1 & 84.1 & 84.0\\
 RandHash & 83.9 & 83.99 & 83.9 & 83.76\\
 BoE & 83.1 & 83.1 & 83.1 & 83.1 \\
 NTSG & 82.5 & 83.7 & 82.8 & 82.4\\
 SIF &  82.3 & 82.6 & 82.9 & 82.2\\
 BoWV  & 81.6 & 81.1 & 81.1 & 80.9\\
 LTSG & 82.8 & 82.4 & 81.8 & 81.8\\
 p-means & 82.0 & 81.9 & 82.0 & 81.6\\
 WTM & 80.9 & 80.3 & 80.3 & 80.0\\
 w2v-LDA & 77.7 & 77.4 & 77.2 & 76.9\\
 ELMo & 74.1 & 74.0 & 74.1 & 73.9\\
 TV+MeanWV & 72.2 & 71.8 & 71.5 & 71.6\\
 MvTM & 72.2 & 71.8 & 71.5 & 71.6\\
 TWE-1 & 81.5 & 81.2 & 80.6 & 80.6\\
 Lda2Vec & 81.3 & 81.4 & 80.4 & 80.5\\
 LDA & 72.2 & 70.8 & 70.7 & 70.0\\
 weight-AvgVec &  81.9 & 81.7 & 81.9 & 81.7\\
 BoW & 79.7 & 79.5 & 79.0 & 79.0\\
 weight-BOC & 71.8 & 71.3 & 71.8 & 71.4\\
 PV-DBoW  & 75.4 & 74.9 & 74.3 & 74.3 \\
 PV-DM  & 72.4 & 72.1 & 71.5 & 71.5 \\ 
 \hline
\end{tabular}
\end{center}
\label{table:1}
\end{table}
\begin{table}[!ht]
\small
\begin{center}
\caption{Performance on multi-label classification on Reuters.}
\vspace{1.0em}
\label{table:2}
\resizebox{.97\columnwidth}{!}{
\begin{tabular}{ c|c|c|c|c|c|c }
\hline
 \bf Model &\bf Prec &\bf Prec & \bf nDCG & \bf Cover. & \bf LRAPS & \bf F1 \\
 &\bf @1 &\bf @5 &\bf @5 &\bf Error &\bf Score \\ 
 \hline
\psif{}  & \bf {94.92} & \bf {37.98} & \bf {50.40} & \bf {6.03} & \bf {93.95} & \bf {82.87}\\
 (Doc2VecC) & & & & &\\
\psif{}  & \bf {94.77} & \bf {37.33} & \bf {49.97} & \bf {6.24} & \bf {93.72}  & \bf {82.41} \\ \hline
BERT (pr) & 93.8 & 37 & 49.6 & 6.3 & 93.1 & 81.9\\
SCDV & 94.20 & 36.98 & 49.55 & 6.48 & 93.30  & 81.75 \\
Doc2VecC & 93.45 & 36.86 & 49.28 & 6.83 & 92.66 & 81.29\\
p-means & 93.29 & 36.65 & 48.95 & 10.8 & 91.72 & 77.81\\
BoWV & 92.90 & 36.14 & 48.55 & 8.16 & 91.46 & 79.16 \\
TWE & 90.91 & 35.49 & 47.54 & 8.16 & 91.46 & 79.16\\
SIF & 90.40 & 35.09 & 47.32 & 8.98 & 88.10 & 76.78\\
PV-DM & 87.54 & 33.24 & 44.21 &  13.2 & 86.21 & 70.24\\ 
PV-DBoW  & 88.78 &  34.51& 46.42 & 11.3 & 87.43 & 73.68\\
AvgVec  & 89.09 & 34.73 & 46.48 & 9.67 & 87.28 & 71.91\\
tfidf AvgVec & 89.33 & 35.04 & 46.83 & 9.42 & 87.90 & 71.97\\ \hline
\end{tabular}}
\end{center}
\end{table}
\begin{figure}[!ht]
\centering
\includegraphics[width=.88\columnwidth]{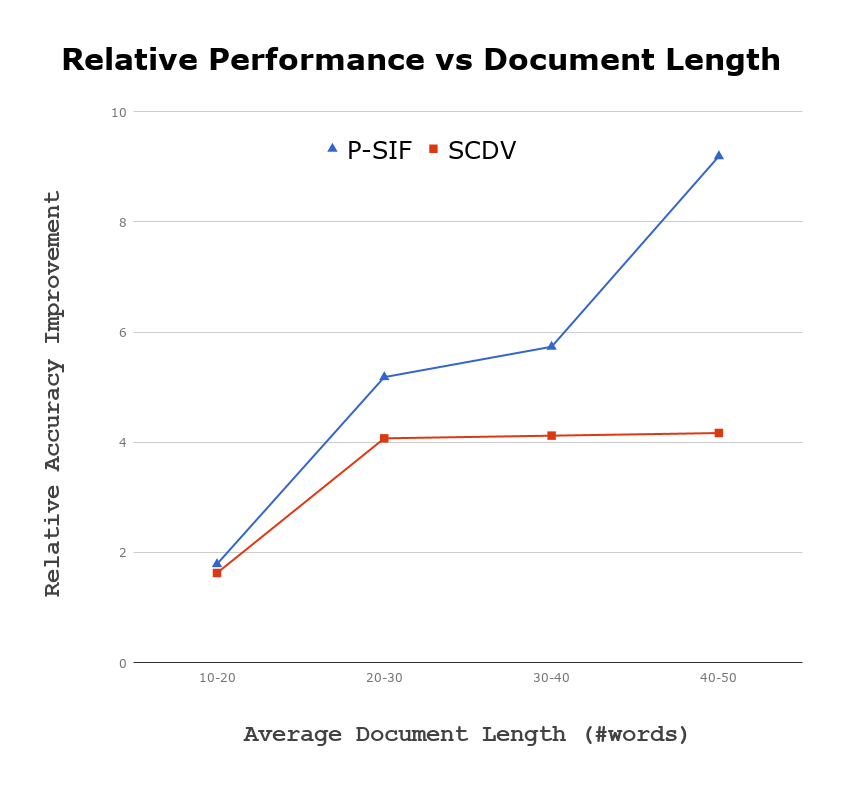}
\caption{Relative performance improvement of \psif{} and SCDV over SIF w.r.t the average document length.}
\label{figure:shortlongnew}
\end{figure}

\textbf{Results and Analysis: }We observe that \psif{} outperforms all other methods by a significant margin on both 20NewsGroup (Table \ref{table:1}) and Reuters (Table \ref{table:2}). We observe that the dictionary learns more diverse and non-redundant topics compared to overlapping clustering (SCDV) since we require only $40$ partitions rather than $60$ partitions in SCDV to obtain the best performance. Simple tf-idf weighted averaging-based document representations do not show significant improvement in performance by increasing word vector dimensions. We achieve a $<0.4\%$ improvement in the accuracy when the word-vector dimensions increase from $200$ to $500$ on 20NewsGroup.  We observe that increasing the word-vectors' dimensions beyond $500$ does not improve  SIF and \psif{}'s performances. We further improve the performance on both datasets using Doc2vecC-initialized \cite{chen2017efficient} word-vectors which reduce word level noise in the \psif{} representations. We represent this approach by \psif{} (Doc2VecC) in Table \ref{table:1} and Table \ref{table:2}. On 20NewsGroup, we require only $20$ partitions instead of $40$ with Doc2VecC-initialized word vectors. This shows that better word vector representations help in learning more diverse and non-redundant partitions. We also report our results (micro-F1) on each of the $20$ classes of 20NewsGroup in the Supplementary material $^5$. Additionally, we empirically show that our proposed embedding \psif{} outperforms the word mover distance \cite{kusner2015word} and performs comparable with the word mover embedding \cite{D18-1482} in Table \ref{tab:my_wnd_comp}. \footnote{For datasets and baseline details refer to \cite{D18-1482}.} Overall, \psif{} outperforms most methods on several datasets by a significant margin. \\

\textbf{Comparison with Contextual Embeddings: } Despite its simplicity, \psif{} is able to outperform unsupervised contextual embeddings such as BERT (pr) and ELMo. We assume the reason behind this is \psif{}'s focused ability to effectively capture both global and local semantics in sparse higher dimension representations. On other hand, BERT tries to capture both syntax and semantics in single lower dimensional continuous representations. In both classification and similarity tasks, understanding syntax is not as prominent as understanding semantics.

\section{Analysis and Discussion}
\textbf{Effect of Document-Length: }We conduct a small experiment to show that our model performs better compared to SIF when we have large size documents. We have divided $26$ STS datasets by average document length, i.e., the number of words in documents in bins of ($10-20$, $20-30$, $30-40$, $40-50$) words. Next, we average the relative performance improvement by \psif{} and SCDV by accuracy with respect to SIF \big($\frac{\text{Method}-\text{SIF}}{\text{SIF}}$\%\big) for datasets in each bin. In Figure \ref{figure:shortlongnew}, we observe that for complex multi-sentence documents with more words, \psif{} performs relatively better than SCDV. We also note that short texts require fewer number of partitions to achieve their best performance which is quite intuitive since short text documents will map into fewer topics.

\textbf{Effect of Sparse Partitioning: }Partitioning and concatenation of word embeddings over topics also helps in the representation of multi-sense words, which would have been left-out by simple averaging of the word embeddings in document representation otherwise. Empirically, on both datasets, we observe that the dictionary learns more diverse and non-redundant topics compared to overlapping clustering because of sparsity constraints. We require only $20$ partitions rather than $60$ in SCDV to obtain the best performance, meaning we just need $(20*300)$ dimensions of embeddings (mostly sparse) compared to $(60*300)$ dimensions of embeddings (mostly non-sparse). Thus, we obtain a performance gain (F1-Score) of $~1.5\%$ with less than $0.33$ of the size of the SCDV embeddings. Lastly, due to fewer dimensions, the feature formation time is less in \psif{}.
\section{Conclusions and Future Work}
We propose a novel unsupervised document feature formation technique based on partitioned word vector averaging. Our embedding retains the simplicity of simple weighted word averaging while taking documents' topical structure into account. Our simple and efficient approach achieves significantly better performance on several textual similarity and textual classification tasks, e.g., we outperform contextual embeddings such as BERT (pr) and ELMo. One limitation of our work is its ignorance of words' order and syntax. In the future, we plan to address this problem and model partitioning, averaging, and learning as a joint process. 
\bibliographystyle{aaai}
\small{
\bibliography{main}}

\appendix

\section{Derivation of \psif{} Embeddings}
\noindent To derive the \psif{} embedding, we propose a generative model which treats corpus generation as a dynamic process where the $t^{th}$ word is produced at step $t$. This process is driven by random walk over a unit norm sphere with the center at the origin. Let $\vec{v}_{c_t}$ be the $d$ dimensional vector from the origin to the current walk point at time $t$. We call this vector the context vector  $\vec{v}_{c_t}$ as it represents the context in the discussion. Let $Z_c$ represent the partition function for the random context vector $\vec{v}_{c_t}$, given by $Z_c = \sum_{w} \exp(\langle\vec{v}_{c_t}, \vec{v}_{w}\rangle)$. $c_0$ and $\vec{v}_{c_0}$ represent a common context and its corresponding $d$ dimensional context vector based on syntax.



\noindent Using log linear model of \cite{mnih2007three}, we define the probability of observing a word $w$ from the random walk with current context $c_t$ at time $t$ as 
\begin{equation}
\small
Pr[w|c_t] \propto \exp(\langle\vec{c}_t, \vec{v}_w\rangle)
\label{eqn:earliermodel}
\end{equation}

\noindent It is easy to show that such random walk under some reasonable assumptions \cite{arora2016latent} can give word-word co-occurrence probabilities similar to empirical works like word2vec \cite{mikolov2013distributed} and GloVe \cite{pennington2014glove}. To account for frequent stop-words which occur more often regardless of context and the common context related to document syntax, two correction terms need be added: one based on $p(w)$ and the other on the common context vector $\vec{v}_{c_{0}}$ in Equation \eqref{eqn:earliermodel}. These terms allow words with a low inner product with $\vec{c}_t$ a chance to appear either from $p(w)$ if they are frequent or by the common context $\vec{c}_0$  if they have a large dot product with $\vec{c}_0$. Given a context vector $c_t$, the probability of a word $w$ in document $d$ being generated by context $c_t$ is given by,
\begin{equation}
\small
Pr[w|c_t] = \lambda p(w) + (1 - \lambda)\frac{\exp(\langle\vec{c}'_t, \vec{v}_w\rangle)}{Z_{c'_t}}
\end{equation}

\noindent where {\small $ \vec{c'_t}$ = $\beta \vec{c}_0 + (1 - \beta)$ $\vec{c}_t$, $\langle\vec{c}_0 , \vec{c}_t \rangle$ $=$ $0$, $\lambda$} and {\small $\beta$} are scalar hyper-parameters. 

\noindent For generating a document from the above random walk-based latent variable model, we consider the following assumptions:

\begin{enumerate}
\itemsep 0.1em 

\item Context vector ($\vec{v}_{c_{0}}$) does not change significantly while words are generated from the random walk, as shown by \cite{arora2016simple}, except the jumps due to topic change.

\item Total number of topics in the entire corpus is $K$, which can be determined by sparse dictionary learning as shown by \cite{arora2016linear} over word vectors $\vec{v}_w$.


\item Word vectors $\vec{v}_w$ are uniformly distributed, thus making the partition function $Z_{c}$ roughly the same in all directions for a given context $c$ emerging from each of the $K$ topics. 
\end{enumerate}

\noindent For a document $d$, the likelihood of document is being generated by the $K$ contexts is given by:
\begin{equation}
\small
 p(d|\{c_1,c_2 \ldots c_K\}) \propto \prod_{j=1}^K{\prod_{\{w \in d\}} p(w|c_{j})}
 \label{eq: 4}
\end{equation}
\begin{equation}
\small
\begin{split}
= \prod_{j=1}^K\prod_{w \in d} \Big[ \lambda p(w) + (1 - \lambda) \frac{\exp(\langle \vec{v}_w, \vec{v}_{{c}_j} \rangle)}{Z_j}  \Big]
 \label{eq: 5}
 \end{split}
\end{equation}
\begin{equation}
\small
\begin{split}
\text{Let,   }  f_w(c_j) = \log\Big[ \lambda p(w)  + (1 - \lambda) \frac{\exp(\langle\vec{v}_w, \vec{v}_{{c}_j}\rangle)}{Z_j}  \Big] 
 \label{eq: 6}
\end{split}
\end{equation}

\noindent Here, {\small $p(w|c_j)$} is the probability that word $w$ is generated by context $c_j$, the value of which is determined by 1) The overall frequency of word $w$ in the corpus, i.e., prior probability ({\small $p(w)$}) and 2) The relative frequency of $w$ appearing with context $j$ with respect to other contexts (determined by {\small $\alpha_{(w,j)}$}). 

\noindent Using simple algebra and treating $p(w)$ as a constant, we can show that $\nabla(f_w(c_j))$ equals, 
\begin{equation}
\small
\begin{split}
\frac{1}{\lambda p(w) + ( 1 - \lambda) \exp(\langle \vec{v}_w, \vec{v}_{c_j}\rangle)/Z_j} * \frac{1 - \lambda}{Z_j}\exp(\langle\vec{v}_w, \vec{v}_{c_j}\rangle)\vec{v}_w
\end{split}
 \label{eq: 7}
\end{equation}

\noindent Then, by using the Taylor expansion, we can show 
\begin{equation}
\small
\begin{split}
f_w(c_j)  \approx f_w(c_j = 0) + \nabla(f_w(c_j = 0))^T\vec{v}_{c_j}
 \label{eq: 8}
 \end{split}
\end{equation}
\begin{equation}
\small
f_w(c_j)  \approx constant + \nabla(f_w(c_j = 0))^T\vec{v}_{c_j}
\end{equation}

\noindent Therefore, the maximum likelihood estimator (MLE) for $\vec{v}_{c_j}$ on the unit sphere (ignoring normalization) given  $a = \frac{1 - \lambda}{\lambda Z_j}$, is approximately \footnote{Note that $argmax_{c:\|\vec{c}\| =1} C + \langle\vec{c},\vec{g}\rangle = \frac{\vec{g}}{\|\vec{g}\|}$ for any constant C}

\begin{equation}
\small
\argmax \sum_{w \in d} f_w(c_j) \propto \sum_{w \in d} \frac{a}{p(w) + a} \vec{v}_w
\end{equation}

\noindent Thus, the MLE estimate is approximately a weighted average of the word-vectors generated from context $j$ in document $d$ from the random walk. We can get the overall context representation $\vec{v}_{c_d}$ of the document by simple concatenation over all $K$ topics i.e. $\vec{v}_{c_d} = \bigoplus_{j=1}^K\vec{v}_{c_j}$.
\noindent Here, $\bigoplus$ represents the concatenation operation. For a document if no word is generated from the context $c_j$ then we can substitute the context vector $\vec{v}_{c_j}$ by a $\vec{0}$ vector to represent $\vec{v}_{c_d}$ in $K \times d$ dimensions. The embedding of a sentence can be obtained by $\vec{v}_{c_s} = \sum_{\{w \in s\}} \frac{a}{p(w) + a}\vec{v}_w$ where $a = \frac{1 - \lambda}{\lambda Z_s}$. \\

\noindent \textbf{Relation to SIF model: } \cite{arora2016simple} show sentences can be represented as averaging of word vectors, under the two assumptions: 
\begin{itemize}
\item uniform distribution of word vectors $\vec{v}_w$  which implies that the partition function $Z_{t}$ is roughly the same in all directions for a sentence.
\item context vector $\vec{v}_{c_h}$ remains constant while the words in the sentence are emitted, implying the replacement of $\vec{v}_{c_h}$ in the sentences by $\vec{v}_{c_s}$ and partition function $Z_t$ by $Z_s$. 
\end{itemize}

\noindent However, the above assumptions do not hold true for a document with multiple sentences where one can expect to have more frequent jumps during a random walk due to topic change. \footnote{It is trivial to assume that these jumps occur more frequently in multi-sentence documents because of more number of topics.} Instead of assuming a single context for the whole document $c_h$, we assume that the total number of contexts over a given corpus is bounded by the number of topics $K$ (as shown by \cite{arora2016linear}), and the random walk can perform jumps to switch context from one context to the rest of $K-1$ contexts.  The partition function remains the same in all directions only for the words emerging from the same context $c_j$ instead of the words coming from all the $K$ contexts. Thus, our approach is a strict generalization of the sentence embedding approach by \cite{arora2016simple} which is a special case of $K = 1$. 

\section{Details of Textual Similarity Task} 
\label{sec:detailresults}
In this supplementary section, we present the details of the STS tasks for each year. Each year, there are $4$ to $6$ STS tasks, as shown in Table \ref{table:stsdatasets}. Note that tasks with the same name in different years are different tasks in reality. We provide detailed results for each tasks in STS 12 - 15 in Table \ref{table:detailresult}. Our method outperforms all other methods from \cite{arora2016simple} and \cite{wieting2015towards} on all $16$ out of $22$ tasks. Our method performs significantly better in comparison to all unsupervised embedding methods. In addition, \psif{} is very close to the best performance by supervised methods on the rest of the datasets. Our method was also able to outperform state of the art supervised averaging based Gated Recurrent Averaging Network (GRAN) \cite{wieting2017revisiting} on $11$ datasets shown in Table \ref{table:detailresult}. Our results also outperform state of the art methods on many recent supervised embedding methods on the STS $16$ task (See  Table \ref{table:sts16resultdetail}).
\begin{table*}[!htbp]
\small
\begin{center}
\caption{The STS tasks by year. Tasks with the same name in different years are different tasks}
\vspace{0.5cm}
\centering
\label{table:stsdatasets}
\begin{tabular}{ c c c c c} 
 \hline
\bf STS12 & \bf STS13 &\bf STS14 &\bf STS15& \bf STS16 \\
 \hline
MSRpar & headline & deft forum & anwsers-forums & headlines\\
MSRvid & OnWN & deft news & answers-students & plagiarism\\
SMT-eur & FNWN & headline & belief & posteditng\\
OnWN & SMT & images & headline & answer-answer\\
SMT-news & & OnWN & images & question-question\\
 & & tweet news & & \\
\hline
\end{tabular}
\end{center}
\end{table*}
\begin{table*}[!htbp]
\begin{center}
\caption{ Experimental results (Pearson's r $\times$ 100) on textual similarity tasks. The highest score in each row is in bold. The methods can be supervised (denoted as Su.), semi-supervised (Se.), or unsupervised (Un.). See the main text for description of the methods. Many results are collected from \cite{wieting2015towards} and \cite{wieting2017revisiting} (GRAN) except the tfidf-GloVe and our representation.}
\small
\centering
\label{table:detailresult}
\begin{tabular}{ c|c|c|c|c|c|c|c|c|c|c|c|c|c|c|c } 
\hline
\multicolumn{1}{p{0.8cm}|}{\centering {\bf TaskType}} & \multicolumn{8}{p{6.0cm}|}{\centering {\bf Supervised}} & \multicolumn{3}{p{2.0cm}|}{\centering {\bf UnSupervised}} & \multicolumn{3}{p{2.0cm}|}{\centering {\bf Semi Supervised}} & \multicolumn{1}{p{0.8cm}}{\centering {\bf \psif{}}}\\
 \hline
{\bf Tasks} & \multicolumn{1}{p{0.2cm}|}{\centering {\bf PP}} & \multicolumn{1}{p{0.6cm}|}{\centering {\bf PP} \\ {\bf proj}} & \multicolumn{1}{p{0.3cm}|}{\centering {\bf DAN}} & \multicolumn{1}{p{0.3cm}|}{\centering {\bf RNN}} & \multicolumn{1}{p{0.8cm}|}{\centering {\bf iRNN}}  & \multicolumn{1}{p{0.8cm}|}{\centering {\bf LSTM} \\ {\bf (no)}}  & \multicolumn{1}{p{0.8cm}|}{\centering {\bf LSTM} \\ {\bf (o.g.)}}  & \multicolumn{1}{p{0.8cm}|}{\centering {\bf GRAN}} & \multicolumn{1}{p{0.2cm}|}{\centering {\bf ST}} & \multicolumn{1}{p{0.2cm}|}{\centering {\bf avg} \\ {\bf Glove}} & \multicolumn{1}{p{0.4cm}|}{\centering {\bf tfidf} \\ {\bf Glove}} & \multicolumn{1}{p{0.3cm}|}{\centering {\bf avg} \\ {\bf PSL}} & \multicolumn{1}{p{0.4cm}|}{\centering {\bf Glove} \\ {\bf +WR }} & \multicolumn{1}{p{0.4cm}|}{\centering {\bf PSL} \\ {\bf +WR}} & \multicolumn{1}{p{0.9cm}}{\centering {\bf\psif}\\ {\bf +PSL}}\\
 \hline

MSRpar & 42.6 & 43.7 & 40.3 & 18.6 & 43.4 & 16.1 & 9.3 & 47.7 & 16.8 & 47.7 & 50.3 & 41.6 & 35.6 & 43.3 &\bf 52.4\\
MSRvid & 74.5 & 74.0 & 70.0 & 66.5 & 73.4 & 71.3 & 71.3 & 85.2 & 41.7 & 63.9 & 77.9 & 60.0 & 83.8 & 84.1 &\bf 85.6\\
SMT-eur & 47.3 & 49.4 & 43.8 & 40.9 & 47.1 & 41.8 & 44.3 & 49.3 & 35.2 & 46.0 & 54.7 & 42.4 & 49.9 & 44.8 &\bf 58.7 \\
OnWN & 70.6 & 70.1 & 65.9 & 63.1 & 70.1 & 65.2 & 56.4 & 71.5 & 29.7 & 55.1 & 64.7 & 63.0 & 66.2 & 71.8 &\bf 72.2 \\
SMT-news & 58.4 &\bf 62.8 & 60.0 & 51.3 & 58.1 & 60.8 & 51.0 & 58.7 & 30.8 & 49.6 & 45.7 & 57.0 & 45.6 & 53.6 & 59.5\\ \hline
STS’12 & 58.7 & 60.0 & 56.0 & 48.1 & 58.4 & 51.0 & 46.4 & 62.5 & 30.8 & 52.5 & 58.7 & 52.8 & 56.2 & 59.5 &\bf 65.7\\
 \hline
headline & 72.4 & 72.6 & 71.2 & 59.5 & 72.8 & 57.4 & 48.5 &\bf 76.1 & 34.6 & 63.8 & 69.2 & 68.8 & 69.2 & 74.1 & 75.7 \\ 
OnWN & 67.7 & 68.0 & 64.1 & 54.6 & 69.4 & 68.5 & 50.4 & 81.4 & 10.0 & 49.0 & 72.9 & 48.0 & 82.8 & 82.0 &\bf 84.4\\
FNWN & 43.9 & 46.8 & 43.1 & 30.9 & 45.3 & 24.7 & 38.4 &\bf 55.6 & 30.4 & 34.2 & 36.6 & 37.9 & 39.4 & 52.4 & 54.8\\
SMT & 39.2 & 39.8 & 38.3 & 33.8 & 39.4 & 30.1 & 28.8 & 40.3 & 24.3 & 22.3 & 29.6 & 31.0 & 37.9 & 38.5 &\bf 41.0\\ \hline
STS’13 & 55.8 & 56.8 & 54.2 & 44.7 & 56.7 & 45.2 & 41.5 & 63.4 & 24.8 & 42.3 & 52.1 & 46.4 & 56.6 & 61.8 &\bf 64.0\\
 \hline
deft forum & 48.7 & 51.1 & 49.0 & 41.5 & 49.0 & 44.2 & 46.1 &\bf 55.7 & 12.9 & 27.1 & 37.5 & 37.2 & 41.2 & 51.4 & 53.2 \\
deft news & 73.1 & 72.2 & 71.7 & 53.7 & 72.4 & 52.8 & 39.1 &\bf 77.1 & 23.5 & 68.0 & 68.7 & 67.0 & 69.4 & 72.6 & 75.2 \\
headline & 69.7 & 70.8 & 69.2 & 57.5 & 70.2 & 57.5 & 50.9 &\bf 72.8 & 37.8 & 59.5 & 63.7 & 65.3 & 64.7 & 70.1 & 70.2\\
images & 78.5 & 78.1 & 76.9 & 67.6 & 78.2 & 68.5 & 62.9 &\bf 85.8 & 51.2 & 61.0 & 72.5 & 62.0 & 82.6 &  84.8 & 84.8\\
OnWN & 78.8 & 79.5 & 75.7 & 67.7 & 78.8 & 76.9 & 61.7 & 85.1 & 23.3 & 58.4 & 75.2 & 61.1 & 82.8 & 84.5 &\bf 88.1\\
tweet news & 76.4 & 75.8 & 74.2 & 58.0 & 76.9 & 58.7 & 48.2 &\bf 78.7 & 39.9 & 51.2 & 65.1 & 64.7 & 70.1 & 77.5 & 77.5\\ \hline
STS’14 & 70.9 & 71.3 & 69.5 & 57.7 & 70.9 & 59.8 & 51.5 &\bf 75.8 & 31.4 & 54.2 & 63.8 & 59.5 & 68.5 & 73.5 & 74.8\\
 \hline
ans-forum & 68.3 & 65.1 & 62.6 & 32.8 & 67.4 & 51.9 & 50.7 &\bf 73.1 & 36.1 & 30.5 & 45.6 & 38.8 & 63.9 & 70.1 & 70.7\\
ans-student & 78.2 & 77.8 & 78.1 & 64.7 & 78.2 & 71.5 & 55.7 & 72.9 & 33.0 & 63.0 & 63.9 & 69.2 & 70.4 & 75.9 & \bf 79.6\\
belief & 76.2 & 75.4 & 72.0 & 51.9 & 75.9 & 61.7 & 52.6 &\bf 78 & 24.6 & 40.5 & 49.5 & 53.2 & 71.8 & 75.3 & 75.3\\
headline & 74.8 & 75.2 & 73.5 & 65.3 & 75.1 & 64.0 & 56.6 &\bf 78.6 & 43.6 & 61.8 & 70.9 & 69.0 & 70.7 & 75.9 & 76.8\\
images & 81.4 & 80.3 & 77.5 & 71.4 & 81.1 & 70.4 & 64.2 &\bf 85.8 & 17.7 & 67.5 & 72.9 & 69.9 & 81.5 & 84.1 & 84.1\\ \hline
STS’15 & 75.8 & 74.8 & 72.7 & 57.2 & 75.6 & 63.9 & 56.0 &\bf 77.7 & 31.0 & 52.7 & 60.6 & 60.0 & 71.7 & 76.3 &\bf 77.3\\
 \hline
SICK’14 & 71.6 & 71.6 & 70.7 & 61.2 & 71.2 & 63.9 & 59.0 & 72.9 & 49.8 & 65.9 & 69.4 & 66.4 & 72.2 & 72.9 &\bf 73.4\\ 
Twitter’15 & 52.9 & 52.8 & 53.7 & 45.1 & 52.9 & 47.6 & 36.1 & 50.2 & 24.7 & 30.3 & 33.8 & 36.3 & 48.0 & 49.0 &\bf 54.9\\
\hline
\end{tabular}
\end{center}
\end{table*}
\begin{table*}[!htbp]
\caption{Experimental results (Pearson's r $\times$ 100) on textual similarity tasks on STS 16. The highest score in each row is in bold. }
\small
\begin{center}
\label{table:sts16resultdetail}
\begin{tabular}{ c|c|c|c|c|c|c|c|c|c|c|c|c }
\hline
{\bf Tasks} & \multicolumn{1}{p{0.9cm}|}{\centering {\bf Skip} \\ {\bf thoughts}} & \multicolumn{1}{p{0.4cm}|}{\centering {\bf LSTM}} & \multicolumn{1}{p{0.6cm}|}{\centering {\bf Tree} {\bf LSTM}} & \multicolumn{1}{p{1.0cm}|}{\centering {\bf Sent2Vec}}  & \multicolumn{1}{p{1.0cm}|}{\centering {\bf Doc2Vec}} & \multicolumn{1}{p{0.6cm}|}{\centering {\bf GloVe} \\ {\bf Avg}}  & \multicolumn{1}{p{0.80cm}|}{\centering {\bf GloVe} \\ {\bf tf-idf}}  & \multicolumn{1}{p{0.60cm}|}{\centering {\bf PSL} \\ {\bf Avg}}  & \multicolumn{1}{p{0.80cm}|}{\centering {\bf PSL} \\ {\bf tf-idf}}& \multicolumn{1}{p{0.60cm}|}{\centering {\bf GloVe} \\ {\bf +WR}} & \multicolumn{1}{p{0.60cm}|}{\centering {\bf PSL} \\ {\bf +WR}} & \multicolumn{1}{p{0.9cm}}{\centering\bf \psif{}\\ {\bf +PSL}}\\
\hline 
headlines & 51.02 & 75.7 & 74.08 & 75.06 & 69.16 & 49.66 & 52.76 & 70.86 & 72.24 & 72.86 & 74.48 &\bf 75.6 \\
plagiarism & 66.71 & 71.73 & 67.62 & 80.06 & 80.6 & 59.84 & 61.48 & 77.96 & 80.06 & 79.46 & 79.74 &\bf 81.6 \\
post editing & 69.95 & 72.31 & 70.65 & 82.85 & 82.85 & 59.89 & 62.34 & 80.41 & 81.45 & 82.03 & 82.05 &\bf 83.7 \\
ans-ans & 28.63 & 44.17 & 52.27 & 57.73 & 41.12 & 19.8 & 22.47 & 38.5 & 41.56 & 58.15 & 59.98 &\bf 60.2 \\
ques-ques & 40.46 & 60.69 & 55.26 & 73.03 &\bf 73.03 & 46.84 & 56.58 & 48.69 & 59.1 & 69.36 & 66.41 & 67.2 \\ \hline
STS16 & 51.4 & 64.9 & 64.0 &\bf 73.7 & 69.4 & 47.2 & 51.1 & 63.3 & 66.9 & 72.4 & 72.5 &\bf 73.7 \\
 \hline
\end{tabular}
\end{center}
\end{table*}
\section{Experimental Details}
\label{sec:parametersettings}
\subsection{Textual Similarity Task: }
\label{sec:parametersettingtst}
We use the {PARAGRAM-SL999 (PSL)} from \cite{wieting2015paraphrase} as word embeddings, obtained by training on the PPDB \cite{ganitkevitch2013ppdb} dataset \footnote{For a fair comparison with SIF  we use PSL vectors instead of unsupervised GloVe and Word2Vec vectors.}. We use the fixed weighting parameter $a$ value of $10^{-3}$, and the word frequencies $p(w)$ are estimated from the common-crawl dataset. We tune the number of contexts $(K)$ to minimize the reconstruction loss over the word-vectors. We fix the non-zero coefficient $k$ = $K/2$, for the SIF experiments. For the GMM-based partitioning of the vocabulary, we tune the number of clusters' parameter $K$ through a 5-fold cross validation.

\begin{enumerate}
\itemsep 0.1em 
\item \textbf{Unsupervised}: We used ST, avg-GloVe, tfidf-GloVe, and GloVe + WR as a baseline. ST denotes the skip-thought vectors by \cite{kiros2015}, avg-GloVe denotes the unweighted average of the GloVe Vectors by \cite{pennington2014glove} \footnote{We used the 300-dimensional word vectors that are publicly available at \url{http://nlp.stanford.edu/projects/glove/}.}, and tfidf-Glove denotes the tf-idf weighted average of GloVe vectors. We also compared our method with the SIF weighting ($W$) common component removal ($R$) GloVe vectors (GloVe + $WR$) by \cite{arora2016simple}. 
For STS 16, we also compared our embedding with Skip-Thoughts \cite{kiros2015}, BERT pretrained embedding average \cite{devlin2018bert} , Universal Sentence Encoder \cite{cer2018universal} and Sent2Vec \cite{pgj2017unsup} embeddings.

\item \textbf{Semi-Supervised}: We used avg-PSL, PSL + WR, and the avg-PSL used the unweighted average of the PARAGRAM-SL999 (PSL) word vectors by \cite{wieting2015paraphrase} as a baseline, obtained by training on PPDB dataset\cite{ganitkevitch2013ppdb}. The word vectors are trained using unlabeled data. Furthermore, sentence embeddings are obtained from unweighted word vectors averaging. We also compared our method with the SIF weighting (W) common component removal (R) PSL word vectors (PSL + WR) by \cite{arora2016simple}.

\item \textbf{Supervised}: We compared our method with PP, PP-proj., DAN, RNN, iRNN, LSTM (o.g), LSTM(no) and GRAN. All these methods are initialized with PSL word vectors and then trained on the PPDB dataset \cite{ganitkevitch2013ppdb}. PP\cite{wieting2015towards} is the average of word vectors, while PP-proj is the average of word vectors followed by a linear projection. The word vectors are updated during the training. DAN denotes the deep averaging network of \cite{iyyer2015deep}. RNN is a Recurrent neural network, iRNN is the identity activated Recurrent Neural Network based on identity-initialized weight matrices. The LSTM is the version from \cite{gers2002learning}, either with output gates (denoted as LSTM (o.g.)) or without (denoted as LSTM (no)). GRAN denotes state of the art supervised averaging based Gated Recurrent Averaging Network from \cite{wieting2017revisiting}. For STS 16 we also compared our embedding with Tree-LSTM \cite{tai2015improved} embedding.
\end{enumerate}

\subsection{Textual Classification Task: }
\label{sec:parametersettingsmulticlass}
We fix the document embeddings and only learn the classifier. We learn word vector embeddings using Skip-Gram with a window size of $10$, Negative Sampling (SGNS) of $10$, and minimum word frequency of $20$. We use $5$-fold cross-validation on the $F1$ score to tune hyperparameters. We use LinearSVM for multi-class classification and Logistic regression with the OneVsRest setting for multi-label classification. We fix the number of dictionary elements to either $40$ or $20$ (with Doc2vecC initialize word vectors) and non-zero coefficient to $k$ = $K/2$ during dictionary learning for all experiments. We use the best parameter settings, as reported in all our baselines to generate their results. We use $200$ dimensions for tf-idf weighted word-vector model, $400$ for paragraph vector model, $80$ topics and $400$ dimensional vectors for TWE, NTSG, LTSG and $60$ topics and $200$ dimensional word vectors for SCDV \cite{mekala2017scdv}.

\textbf{Baseline Details:}
We considered the following baselines: The Bag-of-Words (BoW) model \cite{harris54}, the Bag of  Word Vector (BoWV) \cite{vivek} model, Sparse Composite Document Vector (SCDV) \cite{mekala2017scdv} \footnote{\url{https://github.com/dheeraj7596/SCDV}} paragraph vector models \cite{le2014distributed}, Topical word  embeddings (TWE-1)  \cite{AAAI159314}, Neural Tensor Skip-Gram Model (NTSG-1 to NTSG-3) \cite{liu2015learning}, tf-idf weighted average word-vector model\cite{pranjal2015weighted} and weighted Bag of Concepts (weight-BoC) \cite{boc} where we built document-topic vectors by counting the member words in each topic, and Doc2VecC \cite{chen2017efficient} where averaging and training of word vectors are done jointly. Moreover, we used SIF \cite{arora2016simple} smooth inverse frequency weight with common component removal from weighted average vectors as a baseline. We also compared our results with other topic modeling based document embedding methods such as WTM \cite{wtm}, w2v-LDA \cite{wtvlda}, LDA \cite{liu2014topic}, TV+MeanWV \cite{tvMeanWV}), LTSG \cite{ltsg}, Gaussian-LDA  \cite{gaussianlda}, Topic2Vec \cite{topic2vec}, Lda2Vec \cite{lda2vec}, MvTM \cite{mvtm} and BERT \cite{devlin2018bert}. For BERT, we reported the results on the unsupervised pre-trained (pr) model because of a fair comparison to our approach which is also unsupervised. 

\section{Class wise Performance on 20NewsGroup}
\label{sec:classwise}
We also reported the precision, recall, and micro-F1 results of separate 20 classes of the 20 NewsGroup dataset. We compared our embedding (\psif{}) with Bag of Words, and SCDV embeddings. In Table \ref{table:classwise}, \psif{} (Doc2VecC) (20 partitions) embeddings outperforms SCDV (60 partitions) on $18$ out of the $20$ classes.

\begin{table*}[!ht]
\caption{Class performnce on the 20newsgroup dataset. P-SIF represents our embedding with 40 partitions. \psif{} (Doc2VecC) represents our embeddings initialized with Doc2VecC trained word-vectors with 20 partitions.}
\small
\begin{center}
\begin{tabular}{ c|c|c|c|c|c|c|c|c|c|c|c|c }
\hline
\multicolumn{1}{c|}{} & \multicolumn{3}{c|}{BoW} & \multicolumn{3}{c|}{SCDV} & \multicolumn{3}{c|}{\psif{}}  & \multicolumn{3}{c}{\centering \psif{} (Doc2VecC)}\\
 \hline
 {\bf Class Name} & {\bf Pre.} & {\bf Rec.} & {\bf F-mes} & {\bf Pre.} & {\bf Rec.} & {\bf F-mes} & {\bf Pre.} & {\bf Rec.} & {\bf F-mes} & {\bf Pre.} & {\bf Rec.} & {\bf F-mes} \\
 \hline
alt.atheism & 67.8 & 72.1 & 69.8 & 80.2 & 79.5 & 79.8 & 83.3 & 80.2 &\bf 81.72 & 83 & 79.9 &  81.4 \\
comp.graphics & 67.1 & 73.5 & 70.1 & 75.3 & 77.4 & 76.3 & 76.6 & 78.1 & 77.3 & 76.8 & 79.2 &\bf 77.9 \\
comp.os.ms-windows.misc & 77.1 & 66.5 & 71.4 & 78.6 & 77.2 &\bf 77.8 & 76.3 & 77.7 & 76.9 & 77.2 & 78.2 & 77.7 \\
comp.sys.ibm.pc.hardware & 62.8 & 72.4 & 67.2 & 75.6 & 73.5 &\bf 74.5 & 73.4 & 74.5 & 73.9 & 71.1 & 74.2 & 72.6 \\
comp.sys.mac.hardware & 77.4 & 78.2 & 77.8 & 83.4 & 85.5 & 84.4 & 87.1 & 84.4 & 85.7 & 87.5 & 87.5 &\bf 87.5 \\
comp.windows.x & 83.2 & 73.2 & 77.8 & 87.6 & 78.6 & 82.8 & 89.3 & 78 & 83.2 & 88.8 & 78.5 &\bf 83.3 \\
misc.forsale & 81.3 & 88.2 & 84.6 & 81.4 & 85.9 & 83.5 & 82.7 & 88 & \bf 85.2 & 82.4 & 86.4 & 84.3 \\
rec.autos & 80.7 & 82.8 & 81.7 & 91.2 & 90.6 & 90.9 & 93 & 90.1 & 91.5 & 92.8 & 90.7 & \bf 91.7 \\
rec.motorcycles & 92.3 & 87.9 & 90.0 & 95.4 & 95.7 & 95.5 & 93.6 & 95.5 & 94.5 & 97 & 96.5 &\bf 96.7 \\
rec.sport.baseball & 89.8 & 89.2 & 89.5 & 93.2 & 94.7 & 93.9 & 93.3 & 95.2 & 94.2 & 95.2 & 95.7 &\bf 95.4 \\
rec.sport.hockey & 93.3 & 93.7 & 93.5 & 96.3 & 99.2 & 97.7 & 95.6 & 98.5 & 97.0 & 96.8 & 98.8 &\bf 97.7 \\
sci.crypt & 92.2 & 86.1 & 89.0 & 92.5 & 94.7 & 93.5 & 89.8 & 93.2 & 91.47 & 93.4 & 96.7 & \bf 95.0 \\
sci.electronics & 70.9 & 73.3 & 72.08 & 74.6 & 74.9 & 74.7 & 79.6 & 78.6 &\bf 79.1 & 78 & 79.3 & 78.6 \\
sci.med & 79.3 & 81.3 & 80.2 & 91.3 & 88.4 & 89.8 & 91.9 & 88.6 & 90.2 & 92.7 & 89.9 & \bf 91.2 \\
sci.space & 90.2 & 88.3 & 89.2 & 88.5 & 93.8 & 91.07 & 89.4 & 94 & 91.6 & 90.7 & 94.4 &\bf 92.5 \\
soc.religion.christian & 77.3 & 87.9 & 82.2 & 83.3 & 92.3 & 87.5 & 84 & 94.3 & 88.8 & 86 & 92.5 &\bf 89.1 \\
talk.politics.guns & 71.7 & 85.7 & 78.0 & 72.7 & 90.6 & 80.6 & 73.1 & 91.2 & 81.1 & 77.3 & 89.8 & \bf 83.1 \\
talk.politics.mideast & 91.7 & 76.9 & 83.6 & 96.2 & 95.4 & 95.8 & 97 & 94.5 & 95.7 & 97.5 & 94.2 &\bf 95.8 \\
talk.politics.misc & 71.7 & 56.5 & 63.2 & 80.9 & 59.7 & 68.7 & 81 & 59 & 68.2 & 82 & 62 &\bf 70.6 \\
talk.religion.misc & 63.2 & 55.4 & 59.04 & 73.5 & 57.2 & 64.3 & 72.2 & 59 &\bf 64.9 & 67.4 & 62.4 & 64.8 \\
 \hline
\end{tabular}
\end{center}
\label{table:classwise}
\end{table*}
\begin{table*}[htbp]
\small
\begin{center}
\caption{Results on similarity, entailment, and sentiment tasks. The row for similarity
(SICK) shows Pearson's r $\times$ 100 and the last two rows show accuracy. The highest score in
each row is in bold. Results in Column 2 to 6 are collected from \cite{wieting2015towards}, and
those in Column 7 for skip-thought are from \cite{kiros2015}, Column 8 for PSL + WR are from \cite{arora2016simple}.}
\vspace{0.5cm}
\label{table:sickresult}
\begin{tabular}{ c|c|c|c|c|c|c|c|c } 
 \hline
{\bf Tasks} & \multicolumn{1}{p{0.4cm}|}{\centering {\bf PP}} & \multicolumn{1}{p{0.6cm}|}{\centering {\bf DAN}} & \multicolumn{1}{p{0.6cm}|}{\centering {\bf RNN}} & \multicolumn{1}{p{0.8cm}|}{\centering {\bf LSTM} \\ {\bf (no)}}  & \multicolumn{1}{p{1.1cm}|}{\centering {\bf LSTM} \\ {\bf (o.g.)}} & \multicolumn{1}{p{1.0cm}|}{\centering {\bf skip} \\ {\bf thought}}  & \multicolumn{1}{p{0.8cm}|}{\centering {\bf PSL} \\ {\bf +WR}} & \multicolumn{1}{p{0.9cm}}{\centering {\bf \psif{}}\\ {\bf +PSL}}\\
 \hline
similarity (SICK) & 84.9 & 85.96 & 73.13 & 85.45 & 83.4 & 85.8 & 86.3 &\bf 87.6  \\
entailment (SICK) & 83.1 & 84.5 & 76.4 & 83.2 & 82.0 & - & 84.6 &\bf 85.5\\
sentiment (SST) & 79.4 & 83.4 & 86.5 & 86.6 &\bf 89.2 & - & 82.2 & 86.4\\
\hline
\end{tabular}
\end{center}
\end{table*}
\section{Other Supervised Tasks}
We also considered three out of domain supervised tasks: the SICK similarity task, the SICK entailment task, and the Stanford Sentiment Treebank (SST) binary classification task by \cite{socher2013recursive}. We used the setup similar to \cite{wieting2015towards} and \cite{arora2016simple} for a fair comparison, including the linear projection maps which take the embedding into $2400$ dimensions (same as skip-thought vectors), and is learned during the training. We compared our method to PP, DAN, RNN, LSTM, skip-thoughts and other baselines. Detailed results are in Table \ref{table:sickresult}.

\textbf{Results and Analysis. }Our method (\psif{}) obtains a better performance compared to PSL + WR on all the three tasks similarity, entailment, and sentiment. We obtained the best results for two of the supervised tasks, although many of these methods (DAN, RNN, LSTM) are trained with supervision. Furthermore, the skip thought vectors use a higher dimension of $2400$ instead of $300$ dimensions (which we projected to $2400$ for a fair comparison). Our method wasn't able to outperform the sentiment task compared to supervised tasks because a) due to the antonym problem word-vectors capture the sentimental meaning of words and b) in our weighted average scheme, we didn't assign more weights to sentiment words such as \textit{`not'}, \textit{`good'}, \textit{`bad'}, there may be some important sentiment words which are down-weighted by the SIF weighting scheme. However, we outperform  PSL + WR by a significant margin and have a less performance gap with the best supervised approach.

\section{Proof: Kernels meet Embeddings}

\begin{enumerate}
    \item $K^1(D_A,D_B)$ represents document similarity between the documents represented by average word vectors $d_{x} = \sum_{i} \vec{v}^x_i$ \\ \\
    \textit{Proof: }
    \begin{equation}
        \vec{dv}^A = \frac{1}{n} \sum_{i=1}^{n} \vec{v}_{w^{A}_{i}} 
        \label{eqn:kerneleqn1}
    \end{equation}
    
    \begin{equation}
        \vec{dv}^A = \frac{1}{m} \sum_{j=1}^{m} \vec{v}_{w^{B}_{j}}
        \label{eqn:kerneleqn2}
    \end{equation}
    
    \begin{equation}
    K^1(D_A,D_B) = \langle \vec{dv}^A \cdot \vec{dv}^B \rangle
    \label{eqn:kerneleqn3}
    \end{equation}

    By substituting values from \ref{eqn:kerneleqn1} and \ref{eqn:kerneleqn2} to \ref{eqn:kerneleqn3}, we will get
    \begin{equation}
        K^1(D_A,D_B) =  \langle \frac{1}{n} \sum_{i=1}^{n} \vec{v}_{w^{A}_{i}}  \cdot \frac{1}{m} \sum_{j=1}^{m} \vec{v}_{w^{B}_{j}} \rangle 
    \end{equation}
    
    \begin{equation}
    K^1(D_A,D_B) =  \frac{1}{nm} \langle \sum_{i=1}^{n} \vec{v}_{w^{A}_{i}}  \cdot \sum_{j=1}^{m} \vec{v}_{w^{B}_{j}} \rangle
    \label{eqn:kernelavg}
    \end{equation}

    Using the distributive property of dot product
    
    \begin{equation}
         \langle \sum_{i=1}^{n} \vec{v}_{w^{A}_{i}}  \cdot \sum_{j=1}^{m} \vec{v}_{w^{B}_{j}} \rangle =      \sum_{i=1}^{n} \sum_{j=1}^{m} \langle  \vec{v}_{w^{A}_{i}}  \cdot  \vec{v}_{w^{B}_{j}} \rangle
         \label{eqn:equalityavg}
    \end{equation}
    
    By substituting \ref{eqn:equalityavg} in \ref{eqn:kernelavg}, we will get
    
    \begin{equation}
          K^1(D_A,D_B) =  \frac{1}{nm}\sum_{i=1}^{n} \sum_{j=1}^{m} \langle\vec{v}_{w^{A}_{i}}\cdot \vec{v}_{w^{B}_{j}} \rangle 
    \end{equation}
    
    \item $K^2(D_A,D_B)$ represents the document similarity between the documents represented by topical word vectors \cite{liu2015learning}\\ 
    
    \textit{Proof: }
    \begin{equation}
        \vec{dv}^A = \frac{1}{n} \sum_{i=1}^{n} \vec{tv}_{w^{A}_{i}} = \sum_{i=1}^{n} \alpha_{w^{A}_i} \oplus \vec{v}_{w^{A}_{i}}
        \label{eqn:kerneleqnour11}
    \end{equation}
    
    \begin{equation}
        \vec{dv}^B = \frac{1}{m} \sum_{j=1}^{m} \vec{tv}_{w^{B}_{i}} = \sum_{j=1}^{m} \alpha_{w^{B}_{i}} \oplus \vec{v}_{w^{B}_{i}}
        \label{eqn:kerneleqnour21}
    \end{equation}
    
    \begin{equation}
    K^2(D_A,D_B) = \langle \vec{dv}^A \cdot \vec{dv}^B \rangle
    \label{eqn:kerneleqnour31}
    \end{equation}

    By substituting values from \ref{eqn:kerneleqnour11} and \ref{eqn:kerneleqnour21} to \ref{eqn:kerneleqnour31}, we will get
    
    \begin{equation}
    K^2(D_A,D_B) =  \frac{1}{nm} \langle \sum_{i=1}^{n} \vec{tv}_{w^{A}_{i}}  \cdot \sum_{j=1}^{m} \vec{tv}_{w^{B}_{j}} \rangle 
    \end{equation}
    
    Using the distributive property of dot product
        
    \begin{equation}
        K^2(D_A,D_B) =  \frac{1}{nm}\sum_{i=1}^{n} \sum_{j=1}^{m} \langle\vec{tv}_{w^{A}_{i}}\cdot \vec{tv}_{w^{B}_{j}} \rangle
    \end{equation}
    
    \begin{equation}
    K^2(D_A,D_B) =  \frac{1}{nm}\sum_{i=1}^{n} \sum_{j=1}^{m} \left\langle 
    \big( \alpha_{w^{A}_i} \oplus \vec{v}_{w^{A}_{i}} \big) \cdot \big( \alpha_{w^{B}_{i}} \oplus \vec{v}_{w^{B}_{i}} \big) \right \rangle
    \label{eqn:kernelourfinal}
    \end{equation}
    
    Using the distributive and scalar multiplication property of dot product
    
    \begin{equation}
    \left\langle 
  \big( \alpha_{w^{A}_i} \oplus \vec{v}_{w^{A}_{i}} \big) \cdot \big( \alpha_{w^{B}_{i}} \oplus \vec{v}_{w^{B}_{i}} \big) \right\rangle 
    =   \langle \alpha_{w^{A}} \cdot \alpha_{w^{B}} \rangle  + \langle \vec{v}_{w^{A}_{i}}\cdot \vec{v}_{w^{B}_{i}} \rangle 
    \label{eqn:equalityour14}
    \end{equation}
    
    Since, $ t_{w^{A}}$ =  $\alpha_{w^{A}}$ and $ t_{w^{B}}$ =  $\alpha_{w^{B}}$
    
    \begin{equation}
    \left\langle 
     \big( \alpha_{w^{A}_i} \oplus \vec{v}_{w^{A}_{i}} \big) \cdot \big( \alpha_{w^{B}_{i}} \oplus \vec{v}_{w^{B}_{i}} \big) \right\rangle 
    =   \langle t_{w^{A}} \cdot t_{w^{B}} \rangle  + \langle \vec{v}_{w^{A}_{i}}\cdot \vec{v}_{w^{B}_{i}} \rangle 
    \label{eqn:equalityourfinal}
    \end{equation}
    
    By substituting \ref{eqn:equalityourfinal} in \ref{eqn:kernelourfinal}, we will get
    \begin{equation}
    K^2(D_A,D_B) = \frac{1}{nm} \sum_{i=1}^{n} \sum_{j=1}^{m} \langle \vec{v}_{w^{A}_{i}} \cdot \vec{v}_{w^{B}_{j}}\rangle + \langle\vec{t}_{w^{A}_{i}} \cdot \vec{t}_{w^{B}_{j}}\rangle 
    \end{equation}

    \item $K^3(D_A,D_B)$ represents the document similarity between the documents represented by partition average word vectors (\psif{})\\ 
    
    \textit{Proof: }
    \begin{equation}
        \vec{dv}^A = \frac{1}{n} \sum_{i=1}^{n} \vec{tv}_{w^{A}_{i}} = \sum_{i=1}^{n} \bigoplus_{k=1}^K \alpha_{w^{A}_i,k}\vec{v}_{w^{A}_{i}}
        \label{eqn:kerneleqnour1}
    \end{equation}
    
    \begin{equation}
        \vec{dv}^B = \frac{1}{m} \sum_{j=1}^{m} \vec{tv}_{w^{B}_{i}} = \sum_{j=1}^{m} \bigoplus_{k=1}^K \alpha_{w^{B}_j,k}\vec{v}_{w^{B}_{i}}
        \label{eqn:kerneleqnour2}
    \end{equation}
    
    \begin{equation}
    K^3(D_A,D_B) = \langle \vec{dv}^A \cdot \vec{dv}^B \rangle
    \label{eqn:kerneleqnour3}
    \end{equation}

    By substituting values from \ref{eqn:kerneleqnour1} and \ref{eqn:kerneleqnour2} in \ref{eqn:kerneleqnour3}, we will get
    
    \begin{equation}
    K^3(D_A,D_B) =  \frac{1}{nm} \langle \sum_{i=1}^{n} \vec{tv}_{w^{A}_{i}}  \cdot \sum_{j=1}^{m} \vec{tv}_{w^{B}_{j}} \rangle 
    \end{equation}
    
    Using the distributive property of dot product
        
    \begin{equation}
        K^3(D_A,D_B) =  \frac{1}{nm}\sum_{i=1}^{n} \sum_{j=1}^{m} \langle\vec{tv}_{w^{A}_{i}}\cdot \vec{tv}_{w^{B}_{j}} \rangle
    \end{equation}
    
    \begin{equation}
    \begin{split}
     =  \frac{1}{nm}\sum_{i=1}^{n} \sum_{j=1}^{m} \left\langle 
    \big( \bigoplus_{k=1}^K \alpha_{w^{A}_i,k}\vec{v}_{w^{A}_{i}} \big)\cdot \big( \bigoplus_{k=1}^K \alpha_{w^{B}_j,k}\vec{v}_{w^{B}_{i}} \big) \right \rangle
    \label{eqn:kernelour}
    \end{split}
    \end{equation}
    
    Using distributive and scalar multiplication property of dot product
    
    \begin{equation}
    \begin{split}
         \left\langle 
    \big( \bigoplus_{k=1}^K \alpha_{w^{A}_i,k}\vec{v}_{w^{A}_{i}} \big)\cdot \big( \bigoplus_{k=1}^K \alpha_{w^{B}_j,k}\vec{v}_{w^{B}_{j}}\big) \right\rangle 
    \\ =  \big( \sum_{k=1}^K (\alpha_{w^{A}_i,k} \cdot \alpha_{w^{B}_j,k})\big) \times \langle \vec{v}_{w^{A}_{i}}\cdot \vec{v}_{w^{B}_{i}} \rangle 
    \label{eqn:equalityour1}
    \end{split}
    \end{equation}
    
    Since, $ t_{w^{A},k}$ =  $\alpha_{w^{A},k}$ and $ t_{w^{B},k}$ =  $\alpha_{w^{B},k}$
    
    \begin{equation}
     \begin{split}
    \left\langle 
    \big( \bigoplus_{k=1}^K \alpha_{w^{A},k}\vec{v}_{w^{A}_{i}} \big)\cdot \big( \bigoplus_{k=1}^K \alpha_{w^{B},k}\vec{v}_{w^{B}_{i}}\big) \right\rangle  
    \\=  \big( \sum_{k=1}^K (t_{w^{A}_i,k} \cdot t_{w^{B}_j,k}) \big) \times \langle \vec{v}_{w^{A}_{i}}\cdot \vec{v}_{w^{B}_{i}} \rangle 
    \label{eqn:equalityour2}
    \end{split}
    \end{equation}
    
    \begin{equation}
    \begin{split}
         \left\langle 
    \big( \bigoplus_{k=1}^K \alpha_{w^{A},k}\vec{v}_{w^{A}_{i}} \big)\cdot \big( \bigoplus_{k=1}^K \alpha_{w^{B},k}\vec{v}_{w^{B}_{i}}\big) \right\rangle  
    \\= \big( \sum_{k=1}^K t_{w^{A}_i,k} \cdot t_{w^{B}_j,k} \big) \times \langle \vec{v}_{w^{A}_{i}}\cdot \vec{v}_{w^{B}_{i}} \rangle 
    \label{eqn:equalityour3}
    \end{split}
    \end{equation}
    
    From the definition of the dot product of vectors,
    
    \begin{equation}
    \begin{split}
            \left\langle 
    \big( \bigoplus_{k=1}^K \alpha_{w^{A},k}\vec{v}_{w^{A}_{i}} \big)\cdot \big( \bigoplus_{k=1}^K \alpha_{w^{B},k}\vec{v}_{w^{B}_{i}}\big) \right\rangle  
    \\ =  \langle \vec{t}_{w^{A}_{i}} \cdot \vec{t}_{w^{B}_{j}} \rangle \times   \langle\vec{v}_{w^{A}_{i}} \cdot \vec{v}_{w^{B}_{j}} \rangle
    \end{split}
    \label{eqn:equalityour}
    \end{equation}
    
    By subatituting \ref{eqn:equalityour} in \ref{eqn:kernelour}, we will get
    \begin{equation}
    K^3(D_A,D_B) = \frac{1}{nm} \sum_{i=1}^{n} \sum_{j=1}^{m} \langle \vec{v}_{w^{A}_{i}} \cdot \vec{v}_{w^{B}_{j}}\rangle \times \langle\vec{t}_{w^{A}_{i}} \cdot \vec{t}_{w^{B}_{j}}\rangle 
    \end{equation}

    \item $K^4(D_A,D_B)$ represents the document similarity  between the documents represented by the relaxed word mover distance \cite{kusner2015word} when words of $D_A$ are matched to $D_B$ \\
    \textit{Proof:}
    From the definition of the relaxed word mover distance in \cite{kusner2015word}. Relaxed word mover maps each word $w_i^{A}$ of document $D_A$ to the closest word $w_j^{B}$ of document $D_B$. \\
    
    Since, word $w_j^{B}$ of document $D_B$ is  closer to word  $w_i^{A}$ of document $D_A$, we will have
    
    \begin{equation}
          w_j^{B} = \argmax_{w_j^{B}} \langle \vec{v}_{w^{A}_{i}} \cdot \vec{v}_{w^{B}_{j}} \rangle
    \end{equation}
    
    Therefore, similarity contribution $K_{(i,j)}$ from word $w_i^{A}$ of document $D_A$ is given by:
    \begin{equation}
         K_{(i,j)} = \frac{1}{n} \max_{w_j^{B}} \langle \vec{v}_{w^{A}_{i}} \cdot \vec{v}_{w^{B}_{j}} \rangle
    \end{equation}
    
    Total similarity contribution from all the $n$ words of the document $d_{A}$:
    
    \begin{equation}
         K^4(D_A,D_B) = \frac{1}{n} \sum_{i=1}^n K_{(i,j)} = \frac{1}{n} \sum_{i=1}^n \max_{w_j^{B}} \langle \vec{v}_{w^{A}_{i}} \cdot \vec{v}_{w^{B}_{j}} \rangle
    \end{equation}
    
    We can write $\max_{w_j^{B}}$ as $\max_{j}$, thus finally
    
    \begin{equation}
        K^4(D_A,D_B) = \frac{1}{n} \sum_{i=1}^{n} \max_{j} \langle \vec{v}_{w^{A}_{i}} \cdot \vec{v}_{w^{B}_{j}}\rangle 
    \end{equation}
    
\end{enumerate}

\section{Qualitative Example: Document Similarity}

\begin{figure}[!ht]
\centering
\includegraphics[scale=0.25]{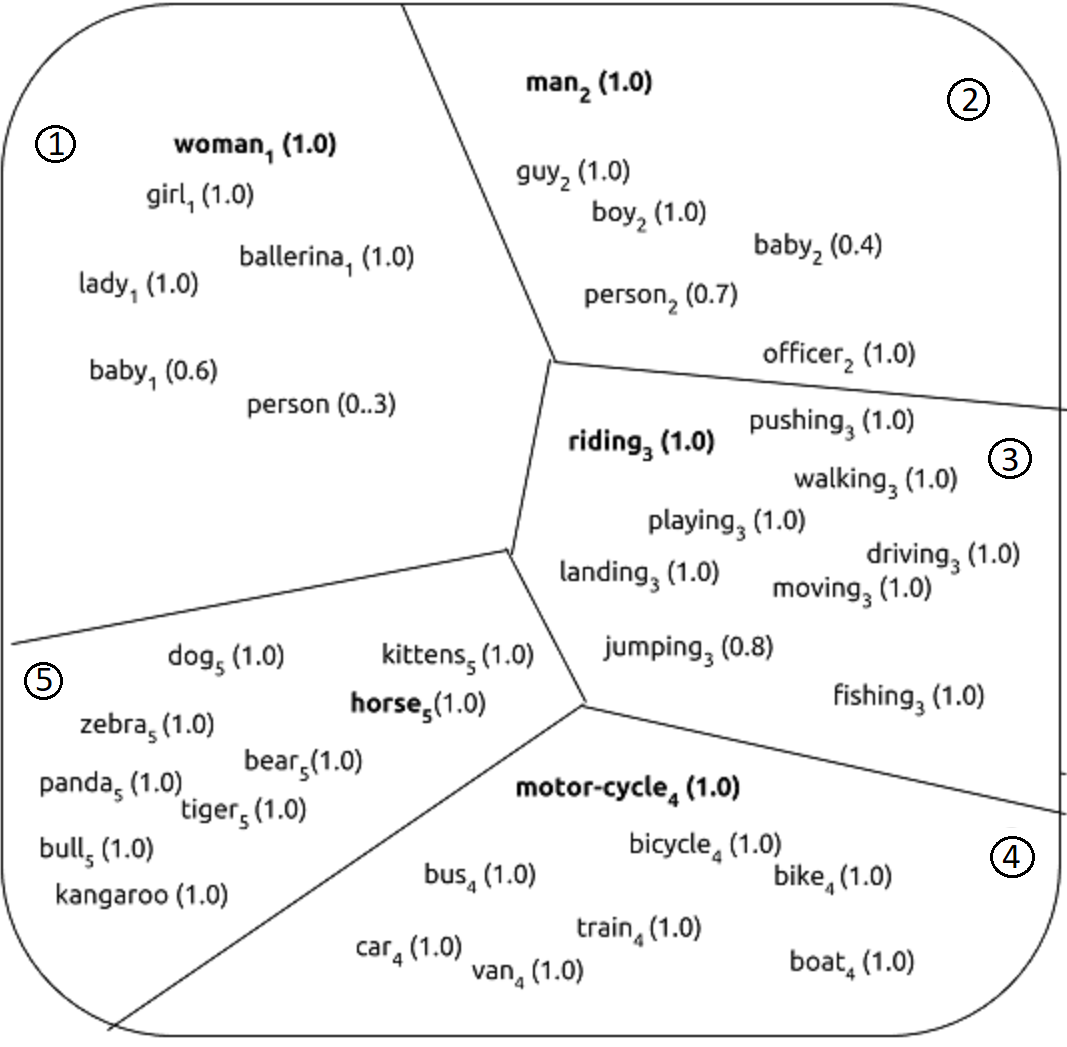}
\caption{Words in different topics are represented by different subscripts  and separated by hyperplanes. Bold represents words from example documents.}
\label{figure:illustration2}
\end{figure}

Let's consider a corpus ($C$) with $N$ documents with the corresponding most frequent vocabulary ($V$). Figure \ref{figure:illustration2} represents the word-vectors space $V$, where similar meaning words are closer. We can apply sparse coding and partition the words-vector space into five (total topics $K=5$) topic vector spaces. Some words are polysemic and belong to multiple topics with some proportion, as shown in Figure \ref{figure:illustration2}. For example, words such as \textit{baby}, \textit{person}, \textit{dog} and \textit{kangaroo}, belong to multiple topics with a significant proportion. Words and corresponding vectors in these topic vector spaces are represented by topic numbers in the subscript. Table \ref{table:twoexam} shows an example pair from the STS Task 2012 MSRVid dataset and the corresponding SIF (averaging) and ~\psif{} (partition averaging) representation vectors. We can see that in the SIF representation, we are averaging words vectors which semantically have different meanings. The document is represented in the same $d$ dimensional word-vectors space. Overall, SIF represents the document as a single point in the vector space and does not take account of different semantic meanings of the topics. Whereas, in the \psif{} representation, we treat the five different semantic topics distinctly. Words belonging to different semantic topics are separated by concatenation ($\oplus$) as they represent different meanings, whereas words coming from the same topic are averaged as they represent the same meaning. The final document vector $\vec{v}_{d_{n}}$ has more representational power as it is represented in a higher 5 $\times$ $d$ dimensional vector space. Thus, partitioned averaging with topic weighting is important for representing documents. 
Empirically, \psif{} assigned a lower score of $0.16$ (rescaled to a 0-1 scale) for sentences ($d_n^{1}$,$d_n^{2}$) where the ground truth is $0.15$ (rescaled to a 0-1 scale), whereas SIF gave similarity score of $0.57$ (0-1 scale), farther than the ground score. Thus, we obtain a relative improvement of $98\%$ in the error difference from the ground truth. Here, the simple averaging-based embedding of $d_n^1$ and $d_n^2$, brings the document representations closer. But partitioned based averaging, \psif{}, projects the documents farther in a higher-dimensional space. 

\begin{table*}[!ht]
\begin{center}
\caption{STS Task 2012 MSRVid dataset similarity example pair. Here, ~\psif{} assigns a score of $0.16$ (rescaled to a 0-1 scale) to sentences ($d_n^{1}$,$d_n^{1}$), where the ground truth of $0.15$ (0-1 scale), whereas SIF assigns a similarity score of $0.57$ (rescaled to a 0-1 scale). Thus, we obtain a relative improvement of $98\%$ in the error difference. Here, $\oplus$ represents concatenation. $\vec{v}_{\mathrm{zero}}$ is the zero padding vector.}
\vspace{0.5em}
\label{table:twoexam}
\begin{tabular}{ c|c|c|c }
\hline
& Document 1 ($d_n^{1}$)  & Document 2 ($d_n^{2}$) & Score \\ 
\hline
Doc & A {$\mathrm{man}$} is {$\mathrm{riding}$} a {$\mathrm{motorcycle}$} &  A {$\mathrm{woman}$} is {$\mathrm{riding}$} a {$\mathrm{horse}$} & 0.15 \\
SIF & $\vec{v}_{\mathrm{man}_2}+ \vec{v}_{\mathrm{riding}_3} + \vec{v}_{\mathrm{motorcycle}_4}$ & $\vec{v}_{\mathrm{woman}_1}+
\vec{v}_{\mathrm{riding}_3} +
\vec{v}_{\mathrm{horse}_5} $ & 0.57\\
P-SIF & $\vec{v}_{\mathrm{zero}_1}$ $\oplus$ $\vec{v}_{\mathrm{man}_2}$ $\oplus$ $\vec{v}_{\mathrm{riding}_3}$ $\oplus$ $\vec{v}_{\mathrm{motorcycle}_4}$ $\oplus$ $\vec{v}_{\mathrm{zero}_5}$ & $\vec{v}_{\mathrm{women}_1}$ $\oplus$ $\vec{v}_{\mathrm{zero}_2}$ $\oplus$ $\vec{v}_{\mathrm{riding}_3}$ $\oplus$ $\vec{v}_{\mathrm{zero}_4}$ $\oplus$ $\vec{v}_{\mathrm{horse}_5}$ & 0.16\\
\hline
\end{tabular}
\end{center}
\end{table*}

\section{Qualitative Results: Similarity task}
Table \ref{table:possExample} represents successful example pair from STS 2012 MSRvid dataset where \psif{} assigns similarity scores closer to ground truth than SIF. Table \ref{table:negExample} represents the failed example pairs from STS 2012 MSRvid dataset where SIF assigns a similarity score closer to the ground truth than \psif{}. We now introduce the header notations used in the Table \ref{table:possExample} and \ref{table:negExample} in details.
\begin{itemize}
\item GT: represents the given ground truth similarity score in a range of 0-5.
\item NGT: represents the normalized ground truth similarity score. NGT is obtained by dividing the GT score by 5 so that it is in a range of 0-1.
\item SIF$_{sc}$: represents the SIF embedding similarity score in a range of 0-1.
\item \psif$_{sc}$: represents the ~\psif{} embedding similarity score in a range of 0-5.
\item SIF$_{err}$: represents absolute error $\| SIF_{sc} - NGT \|$ between normalized ground truth similarity score and the SIF embedding similarity score.
\item \psif$_{err}$: represents the absolute error $\| \psif_{sc} - NGT \|$ between the ground truth similarity score and the ~\psif{} embedding similarity score.
\item Diff$_{err}$: represents absolute difference between SIF$_{err}$ and \psif$_{err}$. Examples where \psif{} performs better Diff$_{err}$ = \psif$_{err}$ - SIF$_{err}$ (used in Table \ref{table:possExample}). Examples where SIF performs better Diff$_{err}$ = SIF$_{err}$ - \psif$_{err}$ (used in Table \ref{table:negExample})
\item Rel$_{err}$: represents relative difference between SIF$_{err}$ and \psif$_{err}$. Examples where \psif{} performs better Rel$_{err}$ = $\frac{Diff_{err}}{SIF_{err}}$ (used in Table \ref{table:possExample}). Examples where SIF performs better Rel$_{err}$ = $\frac{Diff_{err}}{\psif_{err}}$ (used in Table \ref{table:negExample})
\end{itemize}

\begin{table*}
\vspace{-1.0 em}
\caption{STS 2012 MSRVid examples where the ~\psif{} score were far away from the ground truth, whereas the SIF scores were closer to the actual ground truth}
\scriptsize
\begin{center}
\begin{tabular}{ c|c|c|c|c|c|c|c|c|c }
\hline
sentence1 & sentence2 & GT & NGT & SIF$_{sc}$ & ~\psif$_{sc}$ & SIF$_{err}$ &~\psif$_{err}$ &    Diff$_{err}$ & Rel$_{err}$ \\            
\hline
 takes off his sunglasses . & A boy is screaming . & 0.5 & 0.1 & 0.1971 & 0.3944 & 0.0971 & 0.2944 & 0.1973 & 0.6703\\
The rhino grazed on the grass . & A rhino is grazing in a field . & 4 & 0.8 & 0.7275 & 0.538 & 0.0725 & 0.262 & 0.1895 & 0.7234\\
An animal is biting a persons finger . & A slow loris is biting a persons finger . & 3 & 0.6 & 0.6018 & 0.7702 & 0.0018 & 0.1702 & 0.1684 & 0.9892\\
Animals are playing in water . & Two men are playing ping pong . & 0 & 0 & 0.0706 & 0.2238 & 0.0706 & 0.2238 & 0.1532 & 0.6846\\
Someone is feeding a animal . & Someone is playing a piano . & 0 & 0 & -0.0037 & 0.1546 & 0.0037 & 0.1546 & 0.1509 & 0.976\\
The lady sliced a tomatoe . & Someone is cutting a tomato . & 4 & 0.8 & 0.693 & 0.5591 & 0.107 & 0.2409 & 0.1339 & 0.5559\\
The lady peeled the potatoe . & A woman is peeling a potato . & 4.75 & 0.95 & 0.7167 & 0.5925 & 0.2333 & 0.3575 & 0.1242 & 0.3474\\
A man is slicing something . & A man is slicing a bun . & 3 & 0.6 & 0.5976 & 0.4814 & 0.0024 & 0.1186 & 0.1162 & 0.9802\\
A boy is crawling into a dog house . & A boy is playing a wooden flute . & 0.75 & 0.15 & 0.1481 & 0.2674 & 0.0019 & 0.1174 & 0.1155 & 0.9839\\
A man and woman are talking . & A man and woman is eating . & 1.6 & 0.32 & 0.3574 & 0.4711 & 0.0374 & 0.1511 & 0.1137 & 0.7527\\
A man is cutting a potato . & A woman plays an electric guitar . & 0.083 & 0.0166 & -0.1007 & -0.2128 & 0.1173 & 0.2294 & 0.112 & 0.4884\\
A person is cutting a meat . & A person riding a mechanical bull & 0 & 0 & 0.0152 & 0.1242 & 0.0152 & 0.1242 & 0.1091 & 0.8778\\
A woman is playing the flute . & A man is playing the guitar . & 1 & 0.2 & 0.1942 & 0.0876 & 0.0058 & 0.1124 & 0.1065 & 0.948 \\
 \hline
\end{tabular}
\end{center}
\label{table:negExample}
\end{table*}

\begin{table*}
\caption{STS 2012 MSRVid example where the ~\psif{} scores were closer to the ground truth, whereas SIF scores were more away from the ground truth}
\scriptsize
\begin{center}
\begin{tabular}{ c|c|c|c|c|c|c|c|c|c }
\hline
sentence1 & sentence2 & GT & NGT & SIF$_{sc}$ & ~\psif$_{sc}$ & SIF$_{err}$ &~\psif$_{err}$ &    Diff$_{err}$ & Rel$_{err}$ \\            
\hline
People are playing baseball . & The cricket player hit the ball . & 0.5 & 0.1 & 0.2928 & 0.0973 & 0.1928 & 0.0027 & 0.1901 & 0.986 \\
A woman is carrying a boy . & A woman is carrying her baby . & 2.333 & 0.4666 & 0.5743 & 0.4683 & 0.1077 & 0.0017 & 0.106 & 0.9843 \\
A man is riding a motorcycle . & A woman is riding a horse . & 0.75 & 0.15 & 0.5655 & 0.157 & 0.4155 & 0.007 & 0.4085 & 0.9833 \\
A woman slices a lemon . & A man is talking into a microphone . & 0 & 0 & -0.1101 & -0.0027 & 0.1101 & 0.0027 & 0.1074 & 0.9754 \\
A man is hugging someone . & A man is taking a picture . & 0.4 & 0.08 & 0.2021 & 0.0767 & 0.1221 & 0.0033 & 0.1188 & 0.9731 \\
A woman is dancing . & A woman plays the clarinet . & 0.8 & 0.16 & 0.3539 & 0.1653 & 0.1939 & 0.0053 & 0.1886 & 0.9727 \\
A train is moving . & A man is doing yoga . & 0 & 0 & 0.1674 & -0.0051 & 0.1674 & 0.0051 & 0.1623 & 0.9695 \\
Runners race around a track . & Runners compete in a race . & 3.2 & 0.64 & 0.7653 & 0.6438 & 0.1253 & 0.0038 & 0.1214 & 0.9694 \\
A man is driving a car . & A man is riding a horse . & 1.2 & 0.24 & 0.3584 & 0.2443 & 0.1184 & 0.0043 & 0.114 & 0.9636 \\
A man is playing a guitar . & A woman is riding a horse . & 0.5 & 0.1 & -0.0208 & 0.0955 & 0.1208 & 0.0045 & 0.1163 & 0.9629 \\
A man is riding on a horse . & A girl is riding a horse . & 2.6 & 0.52 & 0.6933 & 0.5082 & 0.1733 & 0.0118 & 0.1615 & 0.9319 \\
A woman is deboning a fish . & A man catches a fish . & 1.25 & 0.25 & 0.4538 & 0.2336 & 0.2038 & 0.0164 & 0.1875 & 0.9196 \\
A man is playing a guitar . & A man is eating pasta . & 0.533 & 0.1066 & -0.0158 & 0.0962 & 0.1224 & 0.0104 & 0.112 & 0.915 \\
A woman is dancing . & A man is eating . & 0.143 & 0.0286 & -0.1001 & 0.0412 & 0.1287 & 0.0126 & 0.1161 & 0.9023 \\
The ballerina is dancing . & A man is dancing . & 1.75 & 0.35 & 0.512 & 0.3317 & 0.162 & 0.0183 & 0.1437 & 0.8871 \\
A woman plays the guitar . & A man sings and plays the guitar . & 1.75 & 0.35 & 0.5036 & 0.3683 & 0.1536 & 0.0183 & 0.1353 & 0.8807 \\
A girl is styling her hair . & A girl is brushing her hair . & 2.5 & 0.5 & 0.7192 & 0.5303 & 0.2192 & 0.0303 & 0.1889 & 0.8618 \\
A guy is playing hackysack & A man is playing a key-board . & 1 & 0.2 & 0.3718 & 0.2268 & 0.1718 & 0.0268 & 0.145 & 0.8441 \\
A man is riding a bicycle . & A monkey is riding a bike . & 2 & 0.4 & 0.6891 & 0.4614 & 0.2891 & 0.0614 & 0.2277 & 0.7876 \\
A woman is swimming underwater . & A man is slicing some carrots . & 0 & 0 & -0.2158 & -0.0562 & 0.2158 & 0.0562 & 0.1596 & 0.7397 \\
A plane is landing . & A animated airplane is landing . & 2.8 & 0.56 & 0.801 & 0.6338 & 0.241 & 0.0738 & 0.1672 & 0.6937 \\
The missile exploded . & A rocket exploded . & 3.2 & 0.64 & 0.8157 & 0.6961 & 0.1757 & 0.0561 & 0.1196 & 0.6806 \\
A woman is peeling a potato . & A woman is peeling an apple . & 2 & 0.4 & 0.6938 & 0.5482 & 0.2938 & 0.1482 & 0.1456 & 0.4956 \\
A woman is writing . & A woman is swimming . & 0.5 & 0.1 & 0.3595 & 0.2334 & 0.2595 & 0.1334 & 0.1261 & 0.4859 \\
A man is riding a bike . & A man is riding on a horse . & 2 & 0.4 & 0.6781 & 0.564 & 0.2781 & 0.164 & 0.1142 & 0.4105 \\
A panda is climbing . & A man is climbing a rope . & 1.6 & 0.32 & 0.4274 & 0.3131 & 0.1074 & 0.0069 & 0.1005 & 0.9361 \\
A man is shooting a gun . & A man is spitting . & 0 & 0 & 0.2348 & 0.1305 & 0.2348 & 0.1305 & 0.1043 & 0.444 \\
 \hline
\end{tabular}
\end{center}
\label{table:possExample}
\end{table*}



\end{document}